\documentclass{article}

\usepackage{arxiv}

\usepackage[utf8]{inputenc} 
\usepackage[T1]{fontenc}    
\usepackage{url}            
\usepackage{booktabs}       
\usepackage{amsfonts}       
\usepackage{nicefrac}       
\usepackage{microtype}      
\usepackage{lipsum}
\usepackage{graphicx}
\usepackage{amsmath}
\usepackage{authblk}   
\usepackage{enumitem}  
\usepackage{booktabs}  
\usepackage{caption}   
\usepackage{xcolor}    
\usepackage{subcaption} 
\usepackage{textcomp}
\usepackage{cite}
\usepackage{makecell}

\definecolor{lightgray}{gray}{0.9} 

\usepackage[
colorlinks=true,     
linkcolor=black,     
citecolor=blue,     
urlcolor=blue,       
pdfborder={0 0 0},   
backref=false        
]{hyperref}
\usepackage[nameinlink,capitalise]{cleveref}

\usepackage{easyReview}

\newcommand{\shorttablewidth}{0.48\linewidth} 

\renewcommand{\th}{\textsuperscript{th}} 

\title{HRVConformer: Neonatal Hypoxic-Ischemic Encephalopathy Classification from the Heart Rate signals
	\footnotesize \textsuperscript{}
	\thanks{The work was supported by Taighde Éireann – Research Ireland (19/FFP/6782). This study was also supported by a Strategic Translational Award and an Innovator Award from the Wellcome Trust (098983 \& 209325).}
}

\author[1,2]{Shuwen Yu\thanks{Corresponding author: shuwenyu@umail.ucc.ie}\hspace{0.4em}}
\author[1,2]{William P Marnane}
\author[2,3]{Geraldine B. Boylan}
\author[1,2]{Gordon Lightbody}

\affil[1]{Department of Electrical \& Electronic Engineering, School of Engineering and Architecture\\ University College Cork, Cork, Ireland}
\affil[2]{INFANT Research Centre, University College Cork, Cork, Ireland}
\affil[3]{Pediatrics and Child Health, University College Cork, Cork, Ireland}

\begin{document}
\maketitle

\begin{abstract}
This paper presents the HRVConformer, a novel deep learning architecture for the classification of hypoxic-ischemic encephalopathy (HIE) using the instantaneous heart rate (HR) signal. Unlike conventional approaches that rely on handcrafted features, HRVConformer directly processes raw HR signals in an end-to-end manner, capturing both local and long-range dependencies through a hybrid Convolution-Transformer framework. By integrating convolutional layers for local feature extraction and Transformer-based attention mechanisms for global context modelling, the architecture effectively enhances signal representation and classification performance. The model was trained using supervised learning on a large HR dataset consisting of 1,573 one-hour epochs, including 259 one-hour expert-annotated epochs and a substantial set of weakly labelled data. A 314-hour validation set provided a robust performance estimation, while an independent 215-hour dataset with expert annotations was reserved for final testing. HR signals were extracted from electrocardiogram (ECG) recordings using an improved Pan-Tompkins algorithm, which significantly enhanced both signal quality and data availability. Experimental results demonstrate that the HRVConformer achieves an AUC of 83.23\% and accuracy of 74.56\% on the test set. These results surpass the performance of the Transformer, ResNet50 and fully convolutional networks baselines, highlighting the advantages of integrating convolutional and Transformer-based components for HR-based HIE classification. The proposed method provides a promising step toward a more accurate and automated assessment of HIE using HR signals. The code is available at: \url{https://github.com/syu-kylin/HRVConformer}.
\end{abstract}



\section{Introduction}
Hypoxic-ischemic encephalopathy (HIE) is a severe condition caused by inadequate oxygen and blood flow to the brain in neonates \cite{hankinsDefiningPathogenesisPathophysiology2003,kurinczukEpidemiologyNeonatalEncephalopathy2010}. It carries a substantial burden of morbidity and a high mortality rate \cite{allenHypoxicIschemicEncephalopathy2011,lawnTwoMillionIntrapartumrelated2009}, underscoring the critical need for early and accurate diagnosis. Therapeutic hypothermia (TH) has become a standard care for moderate to server HIE, which cools the infant to the body temperature of 32-34~\textcelsius{} for 72\,h \cite{committee2014hypothermia}. To be effective, this treatment should be initiated within 6 hours after birth \cite{ashooriMachineLearningModels2024, kosteczkoTherapeuticHypothermiaForm2025}. EEG is a gold standard for assessing brain injury in HIE \cite{vergalesDepressedHeartRate2014}, however, it may not be readily available in all clinical settings. Heart rate variability (HRV), a measure of the variation in time between heartbeats, has emerged as a promising biomarker for assessing the severity and outcomes of HIE \cite{addisonHeartRateCharacteristics2009, andersenSeverityHypoxicIschemic2019, gouldingHeartRateVariability2017, bersaniHeartRateVariability2021, massaroEffectTemperatureHeart2017}. Compared with EEG, heart rate variability is much easier to acquire and instantaneously accessible, well suited for rapid making decisions within this narrow window. Recent studies suggest that integrating heart rate variability analysis into clinical practice could facilitate more personalized treatment plans, ultimately contributing to better long-term developmental outcomes for infants suffering from hypoxic-ischemic encephalopathy \cite{vergalesDepressedHeartRate2014, addisonHeartRateCharacteristics2009, edoigiawerieSystematicReviewEEG2024}.  

Heart rate is a non-invasive measure that reflects the function of autonomic nervous system (ANS). It is derived from the intervals between consecutive heartbeats, known as R-R intervals, which can be extracted from electrocardiogram (ECG) signals. The Pan-Tompkins algorithm is a widely used method for detecting these intervals due to its efficiency and accuracy in identifying QRS complexes, even in noisy signals \cite{farihaAnalysisPanTompkinsAlgorithm2020}. It has undergone several improvements to enhance its robustness and performance. Pan-Tompkins++ \cite{imtiazPanTompkinsRobustApproach2022} introduced a moving average filter and multiple sets of thresholds to reduce false positives and false negatives. It has been tested on multiple datasets, showing a 2.8\% reduction in false positives and a 1.8\% reduction in false negatives. Some improvements focused on speed up of the algorithm either quantified the filtering signal \cite{hamiltonQuantitativeInvestigationQRS1986} or removing the background noise \cite{ranaCardiacDiseaseDetection2019}. Hamilton \& Tompkins algorithm \cite{hamiltonQuantitativeInvestigationQRS1986} also utilize mean or median estimators to replace the decision rules. The Pan-Tompkins algorithm has been compared with other peak detection methods, such as the Hilbert Double Envelope Method (HDEM), which shows higher sensitivity compared with Pan-Tompkins algorithm \cite{esgalhadoPeakDetectionHRV2022}. Though some other peak detection methods were proposed, such as wavelet-based methods \cite{sahambiNewApproachOnline1996, goyalStudyHRVDynamics2012, gautamFeatureExtractionHRV2017, vermaRobustAlgorithmDerivation2013} and phasor transformation method \cite{martinezApplicationPhasorTransform2010}, the Pan-Tompkins algorithm is the most popular method due to its simplicity and efficiency.

HRV is influenced by both sympathetic and parasympathetic branches of the ANS, making it a sensitive indicator of neurological and physiological stress. After extraction from the ECG signal, HRV usually can be analysed in three domains: time domain, frequency domain and complexity domain. Features in time domain include metrics such as the mean NN interval (mean NN), standard deviation of NN intervals (SDNN), and root mean square of successive differences (RMSSD). These simple metrics can provide some insights into overall HRV. Spectral analysis in frequency domain decomposes the HRV into different frequency bands, such as low-frequency (LF), high-frequency (HF), and very low-frequency (VLF) components. The LF/HF ratio is often used to assess the balance between sympathetic and parasympathetic activity. Some measures like detrended fluctuation analysis (DFA) and Poincaré plots provide information about the complexity and fractal properties of HRV, which are associated with the integrity of the ANS \cite{gouldingHeartRateVariability2017, pavelHeartRateVariability2023, metzlerPatternBrainInjury2017}.  HRV analysis has a wide range of applications in both clinical and non-clinical settings: cardiovascular disease diagnosis such as arrhythmias \cite{ersoyFeatureExtractionBased2016}, hypertension \cite{turnipDetectionPotentialHypertension2024, semenovaPredictionShorttermHealth2019} and brain injury \cite{yasovabarbeauHeartRateVariability2019, al-shargabiInflammatoryCytokineResponse2017}, and some non-clinical applications like stress detection \cite{wangECGStressDetection2023}, sleep apnea detection \cite{sharanECGDerivedHeartRate2020}.

Studies have shown that HRV is significantly reduced in neonates with HIE compared to healthy controls. The degree of HRV suppression correlates with the severity of brain injury, as assessed by EEG and MRI. For instance, neonates with moderate to severe HIE exhibit lower HRV metrics, such as SDNN, RMSSD, and HF power, compared to those with mild or no injury \cite{metzlerPatternBrainInjury2017, gouldingHeartRateVariability2015}. The LF/HF ratio, an indicator of sympathetic dominance, is lower in infants with severe HIE, suggesting increased stress and impaired parasympathetic activity \cite{aliefendiogluHeartRateVariability2012, andersenSeverityHypoxicIschemic2019}. Combining HRV features and clinical parameters, a logistic regression analysis was implemented to predict EEG grades in neonatal HIE, achieving an AUC of 0.895 on training set \cite{pavelHeartRateVariability2023}, which has confirmed a strong correlation between HRV and HIE grades. Although some clinical and statistical studies have found this relevance between HRV and HIE, it still does not have a standardized deep learning method to verify this.

Deep learning models have demonstrated significant potential in signal processing, offering advanced capabilities for classification, detection, and enhancement across a wide variety of signal types. These models excel at automatically extracting and learning features directly from raw data, which is especially beneficial for handling complex and non-stationary signals. For example, \cite{osathitpornRRWaveNetCompactEndtoEnd2023} proposed RRWaveNet, a multi-scale convolution and residual convolution neural network (CNN), using raw photoplethysmography (PPG) signal as input to extract respiratory rate (RR). \cite{songHeartRateEstimation2020} and \cite{karhadeTimeFrequencyDomain2022} respectively transformed the remote photoplethysmography (rPPG) and phonocardiogram (PCG) signal into 2D spatiotemporal and time-frequency feature image, using deep CNNs for the heart rate extraction and heart valve disorders (HVDs) detection. Combined CNN and bidirectional Long Short-Term Memory, \cite{ramalakshmiDevelopingDeepLearning2024} proposed a CNN-BLSTM model to detect abnormal heartbeats with ECG signal, which outperforms ResNet-50 \cite{he2016deep}and AlexNet \cite{krizhevsky2012imagenet}. The success of Transformer architecture in natural language processing and vision fields, exemplified by the Vision Transformer \cite{dosovitskiyImageWorth16x162021}, has inspired their application in various other fields. For instance, multi-channel Transformer networks have been applied to medical time series data \cite{wang2024medformer, anwarTransformersBiosignalAnalysis2025}. In an effort to capture both local and long-range features, hybrid architectures combining CNNs and Transformers have been used in vision tasks \cite{pengConformerLocalFeatures2021}, speech signals \cite{fuUformerUnetBased2022}, brain computer interface with EEG \cite{songEEGConformerConvolutional2023} and brain injury detection \cite{liCNNInformerHybridDeep2025}.

The convolution-augmented Transformer --- Conformer network, which originally was proposed in \cite{gulatiConformerConvolutionaugmentedTransformer2020}, integrates a convolutional module into the Transformer block to formulate a new Conformer block, which has achieved state-of-the-art accuracy for speech recognition. Some improvements were made to enhance its efficiency by either adding additional encoder-decoder module or employing limited context attention \cite{rekeshFastConformerLinearly2023, kimSEConformerTimeDomainSpeech2021}. However, the performance of this architecture in biomedical signals processing has not been demonstrated. In this paper, inspired by the Vision Transformer and the Conformer architecture, the HRVConformer was proposed here for HIE classification. Large amount of weakly labelled HR data mixed with expert-annotated strongly label data were used for the weakly supervised training of the HRVConformer. From the authors' knowledge, this is the first study to classify HIE using deep learning method directly from the instantaneous heart rate signal. An enhanced version of Pan-Tompkins algorithm is proposed here to improve the quality and quantity of heart rate signal from noisy ECG signals in neonates.

This paper is structured as follows: Section 2 describes the dataset, preprocessing and post-processing procedures, and the enhanced Pan-Tompkins algorithm. Additionally, an overview of the HRVConformer architecture is presented. Section 3 details the experiments and results, including an evaluation of the improved Pan-Tompkins algorithm, a baseline comparison of HRVConformer, and ablation studies. This is followed by a discussion on model performance, attention visualization, and limitations.

\section{Methods}
\subsection{Dataset}
This study is a secondary analysis of data from newborns recruited between January 2011 and February 2017 as part of two multicentre cohort studies conducted across eight European tertiary neonatal intensive care units. The original studies, known as ANSeR1 and ANSeR2, included infants born at or after 36+0 weeks of gestation who required EEG monitoring for clinical reasons. The primary findings have been published, and detailed study protocols are available on \href{https://clinicaltrials.gov/}{ClinicalTrials.gov} (Identifiers: NCT02160171 \cite{rennieCharacterisationNeonatalSeizures2019}, NCT02431780 \cite{pavelMachinelearningAlgorithmNeonatal2020}). Of the 472 neonates enrolled, 284 infants were diagnosed with hypoxic-ischemic encephalopathy (HIE).

Of the 284 neonates with HIE, 35 were excluded either because of a combined diagnosis or because the EEG did not start within the 48 hours after birth; of those 68 were set aside for future validation. This left 181 infants for analysis, with 91 recordings from six centres in the ANSeR1 dataset and 90 recordings from eight centres in the ANSeR2 dataset. All neonates underwent continuous EEG (cEEG) monitoring using either the Nihon Kohden Neurofax EEG-1200 (Tokyo, Japan) or the NicoletOne ICU Monitor \& Xltek (Natus, Middleton, WI, USA), with a sampling frequency of either 256 or 250\,Hz. Due to variations in recording centres, equipment, and protocols, ECG channels were available for only 58 neonates in ANSeR1 and 75 in ANSeR2. 
One-hour epochs were extracted at postnatal ages of 6\th{}, 12\th{}, 24\th{}, 36\th{}, and 48\th{} hours when available. A neurophysiologist annotated each epoch’s hypoxic-ischemic encephalopathy (HIE) grade based on the multi-channel EEG signals. HIE severity was classified into five grades, according to the grading scheme in \cite{murrayEarlyEEGFindings2009}: normal, mild abnormal, moderate abnormal, severely abnormal, and inactive.

\begin{figure}[htbp]
	\centering
	\begin{subfigure}[b]{0.45\textwidth}
			\includegraphics[width=\textwidth]{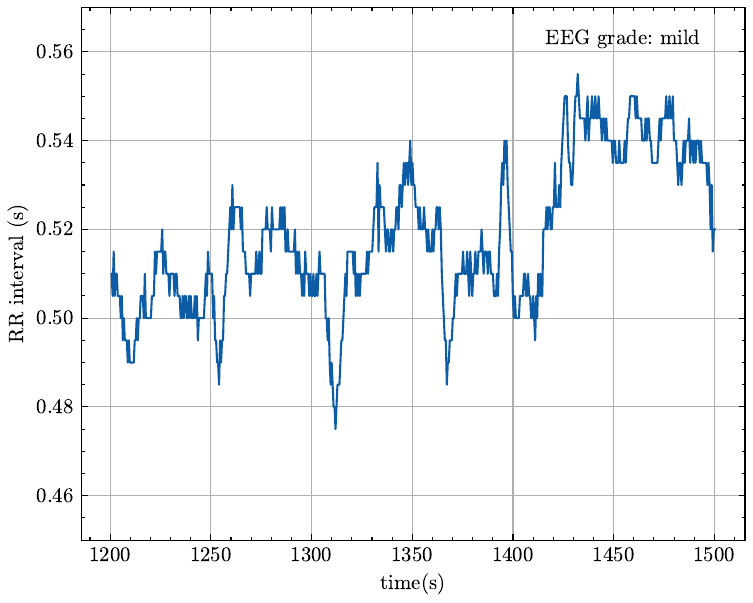}
			\caption{}
			\label{mild_class_rr}
		\end{subfigure}
	\hfill
	\begin{subfigure}[b]{0.45\textwidth}
			\includegraphics[width=\textwidth]{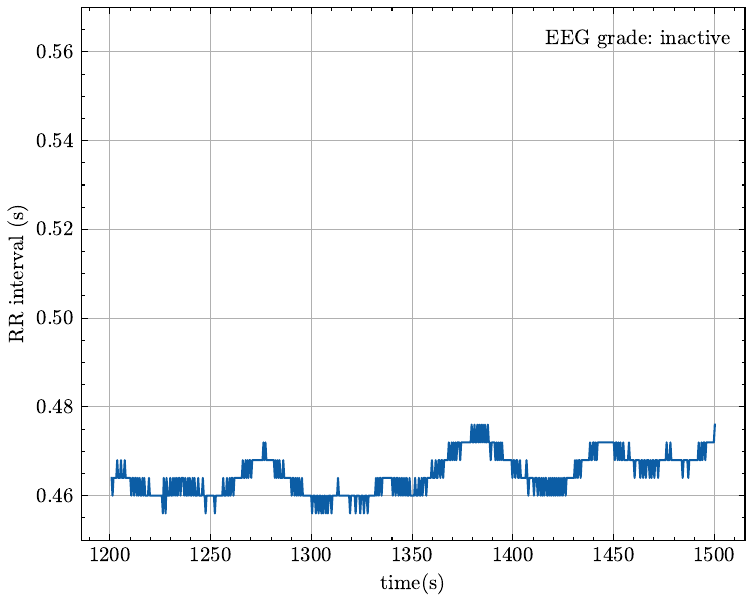}
			\caption{}
			\label{sever_class_rr}
		\end{subfigure}
	\caption{Examples (from the ANSeR dataset) of 5-min RR intervals from different categories. \textbf{(a)} RR interval example from class 0 corresponding to an EEG grade of mild. \textbf{(b)} RR interval from class 1 with regard to an EEG grade of inactive.}
	\label{rr_with_different_grades}
\end{figure}

For this study, this expert-annotated dataset is categorized as strong label group, where normal and mild grades --- no TH, were combined as class 0, and the more severe categories (moderate, severe, and inactive) --- TH, were grouped as class 1. This classification allowed us to generate weak labels for additional one-hour epochs between the 6\th{} and 48\th{} hours. For example, epochs from the 7\th{} to 11\th{} hour would be labelled as class 0 if both the 6\th{} and 12\th-hour epochs are graded as normal or mild. If both epochs have different categories, which means  the intermediate epoch labels can not be determined, thereby they were excluded. The ANSeR2 strong and weak label groups are used for training and validation, while the separate ANSeR1 strong label group is for testing. \cref{rr_with_different_grades} illustrates 5-min RR interval examples from both classes, which can be seen the more severe case shows a less variance and lower amplitude.

\subsection{HR extraction with improved version of Pan-Tompkins algorithm}
The Pan-Tompkins algorithm, originally proposed in \cite{panRealTimeQRSDetection1985}, is a well-established method for extracting heart rate (HR) data from ECG signals. However, as the ANSeR study primarily focuses on EEG signals, less emphasis was placed on the ECG signal quality during data collection. As a result, the standard Pan-Tompkins algorithm struggled to process these signals effectively and, in some cases, failed to extract reliable RR intervals. To address this, an enhanced version of Pan-Tompkins algorithm is designed to improve the quality of extracted RR intervals from noisy and artefact-prone ECG signals that are typical in a clinical settings. The modifications are applied at three key stages: preprocessing, decision-making, and post-processing. The code is available at: \url{https://github.com/syu-kylin/enhanced-Pan-Tompkin}.

\subsubsection{Improvements on preprocessing}
In the preprocessing stage, the ECG signal undergoes a series of transformations, including bandpass filtering, differentiation, squaring, and moving window integration, to extract key features such as slope, amplitude, and width of potential QRS complexes \cite{alvarezComparisonThreeQRS2013}. Modifications in this step include a polarity check to ensure consistent waveform orientation and an optimized frequency band adjustment to enhance QRS detection.

\begin{figure}[htbp]
	\centering
	\begin{subfigure}[b]{0.45\textwidth}
		\includegraphics[width=\textwidth]{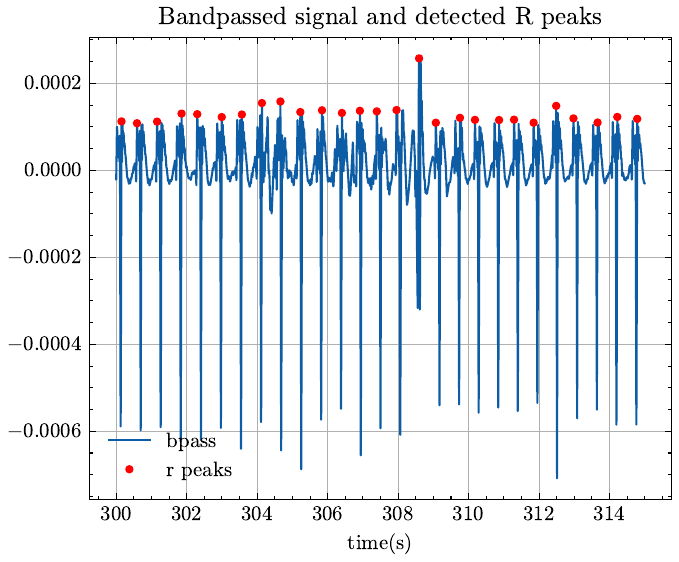}
		\caption{}
		\label{no_reverse_bpass}
	\end{subfigure}
	\hfill
	\begin{subfigure}[b]{0.45\textwidth}
		\includegraphics[width=\textwidth]{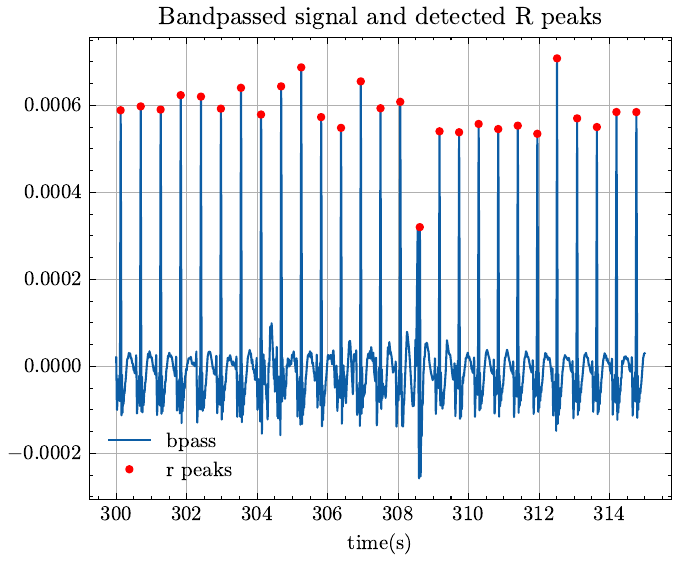}
		\caption{}
		\label{reversed_bpass}
	\end{subfigure}
	\caption{An example of a bandpass filtered ECG signal and detected R peaks \textbf{(a)} without and \textbf{(b)} with polarity correction.}
	\label{bpass_polarity_comparison}
\end{figure}

\begin{figure}
	\centering
	\begin{subfigure}[b]{0.45\textwidth}
		\includegraphics[width=\textwidth]{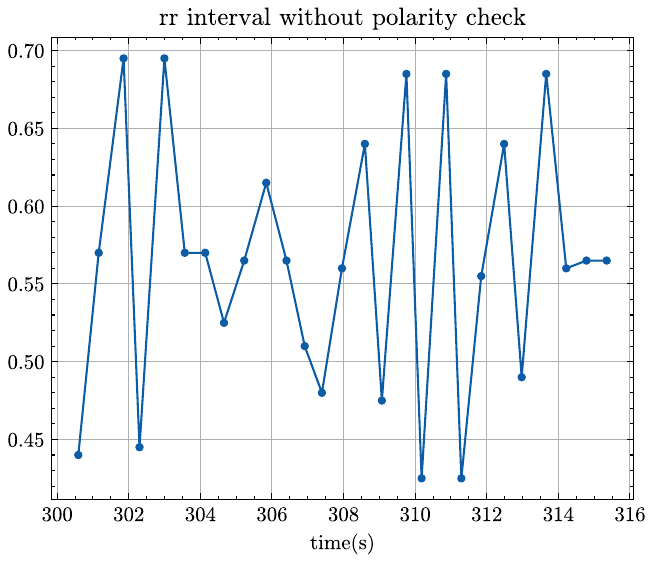}
		\caption{}
		\label{rr_no_reverse}
	\end{subfigure}
	\hfil
	\begin{subfigure}[b]{0.45\textwidth}
		\includegraphics[width=\textwidth]{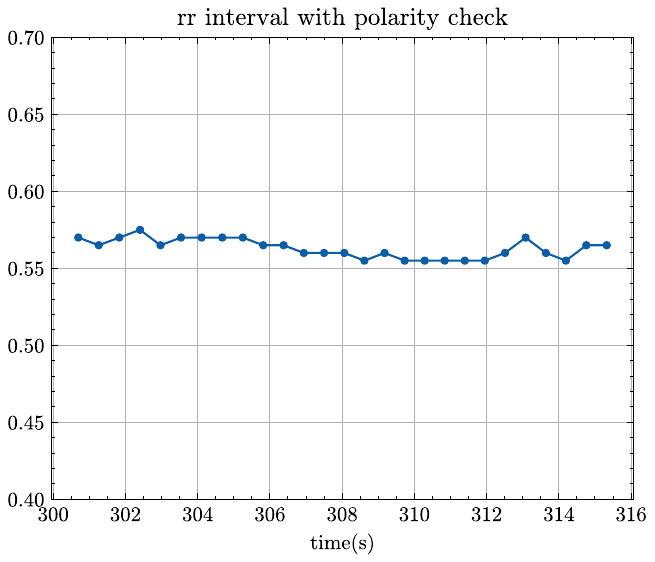}
		\caption{}
		\label{rr_reversed}
	\end{subfigure}
	\caption{Examples of extracted RR intervals \textbf{(a)} without polarity check and \textbf{(b)} with polarity check.}
	\label{rr_polarity_comparison}
\end{figure}

\textbf{Polarity check:} In this work, ECG signal is processed either in a reference montage or bipolar montage, and in some cases, the polarity of selected ECG channel may be reversed. This inversion can lead to inaccuracies in R-peak detection, as illustrated in \cref{bpass_polarity_comparison}. Misidentified R peaks result in highly noisy RR intervals, as shown in \cref{no_reverse_bpass}. After applying polarity correction, the accuracy of QRS complex detection improves significantly, seen in \cref{reversed_bpass}. The resultant RR intervals are demonstrated in \cref{rr_polarity_comparison}, where the detected RR intervals after polarity check are more accurate and less noisy.

\begin{figure}[htbp]
	\centering
	\begin{subfigure}[b]{0.45\textwidth}
		\includegraphics[width=\textwidth]{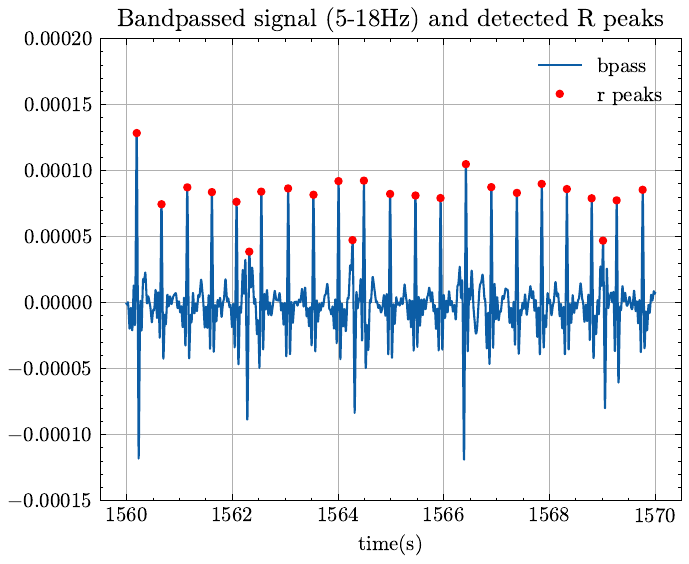}
		\caption{}
		\label{bpass_5-18Hz}
	\end{subfigure}
	\hfill
	\begin{subfigure}[b]{0.45\textwidth}
		\includegraphics[width=\textwidth]{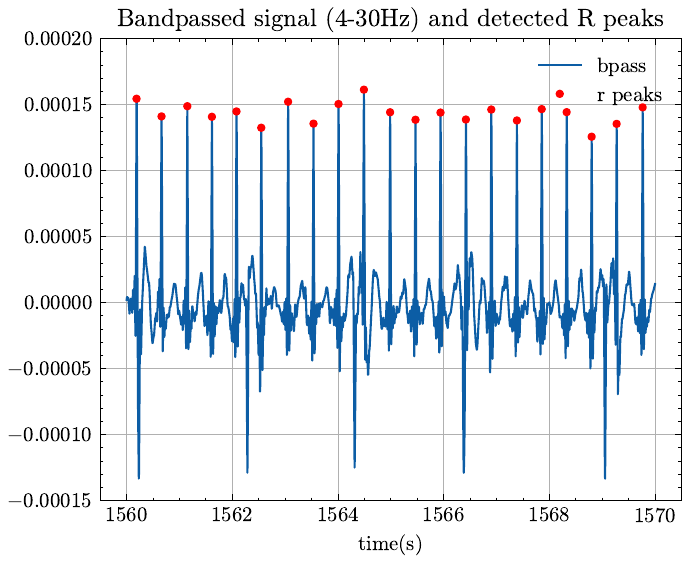}
		\caption{}
		\label{bpass_4-30Hz}
	\end{subfigure}
	\caption{Different frequency bands filtered ECG signal and detected R peaks comparison. \textbf{(a)} 5--18\,Hz bandpass filtered ECG signal and detected R peaks. \textbf{(b)} 4--30\,Hz bandpass filtered ECG signal and detected R peaks.}
	\label{freq_band_comparison}
\end{figure}

\textbf{Frequency Band Extensions:} Recorded ECG signals are often contaminated by various sources of noise and artefacts, including power line interference, baseline wander, motion artefacts, and muscle noise \cite{farihaAnalysisPanTompkinsAlgorithm2020}. These interferences can significantly impact the accurate detection of ECG features. The standard Pan-Tompkins algorithm employs a 5–15\,Hz bandpass filter to mitigate these artefacts; however, this frequency range may exclude important signal components. The Pan-Tompkins++ algorithm \cite{imtiazPanTompkinsRobustApproach2022} suggests using a 5–18\,Hz bandpass filter, which better preserves key frequency components. For infants, relevant frequency components extend further, up to 4–30\,Hz \cite{semenovaPredictionShorttermHealth2019}. Since the ANSeR dataset includes neonates with gestational ages ranging from 36 to 44 weeks, a 4–30\,Hz bandpass filter is more appropriate. \cref{freq_band_comparison} presents a 10-second ECG segment filtered within this range, along with the detected R-peaks. Compared to the 5–18\,Hz filter, the 4–30\,Hz filter retains more critical information, leading to improved QRS complex detection accuracy.

\subsubsection{Improvements on decision-making}
In the decision stage, the Pan-Tompkins algorithm employs a dual-threshold approach to classify potential R-peaks as either signal or noise peaks. These thresholds dynamically adjust based on the moving integration signal and bandpass-filtered signal. However, this method may still fail to detect some low-amplitude R-peaks, thus a search-back procedure is applied. Since R-peak detection is based on the moving integral and bandpass filtered signal, the identified peaks may exhibit a temporal shift relative to the raw ECG signal. To correct this, an additional search back mechanism is implemented to accurately reposition the detected R-peaks on the raw ECG trace. Improvements in this stage include optimized threshold initialization and reset, exclusion of zero-value ECG segments, and an enhanced search back process using the bandpass-filtered ECG signal.

\textbf{Threshold Initialization and Reset:} Before making peak detection decisions, the Pan-Tompkins algorithm requires an initial two-second learning period to analyse ECG signal characteristics and set appropriate thresholds. The threshold initialization method used in this study is adopted from Pan-Tompkins++ \cite{imtiazPanTompkinsRobustApproach2022}:

\begin{align}
I_1 = \frac{Max_I}{3} \\
I_2 = 0.5\,Mean_I \\
S_{\mathrm{PKI}} = I_1 \\
N_{\mathrm{PKI}} = I_2 
\end{align}
where $I_1$ and $I_2$ are the signal and noise threshold for the moving integration signal; $S_{\mathrm{PKI}}$ and $N_{\mathrm{PKI}}$ are the mean estimation of signal and noise peak of the moving integration signal; $Max_I$ and $Mean_I$ represent the maximum and mean amplitude of the 2-second moving integration signal. During initialization, thresholds are set based on the first two seconds of the signal. A second set of thresholds is initialized similarly but works on the bandpass-filtered ECG signal. For the reset procedure, the threshold window is centred at the current time point and extends one second in both directions.

In the standard Pan-Tompkins algorithm, excessively high spikes can cause threshold values to rise too much, preventing the algorithm from properly detecting subsequent R-peaks. To address this issue, a threshold reset is implemented before the search-back step. If no RR intervals are detected for more than 1.4 seconds (considering that the 95\th{} percentile of RR intervals in this dataset is approximately 0.7 seconds, and 1.4 seconds represents an approximate upper limit of two beats) the thresholds are reset to ensure continued peak detection.

\begin{figure}[htbp]
	\centering
	\begin{subfigure}[b]{0.45\textwidth}
		\includegraphics[width=\textwidth]{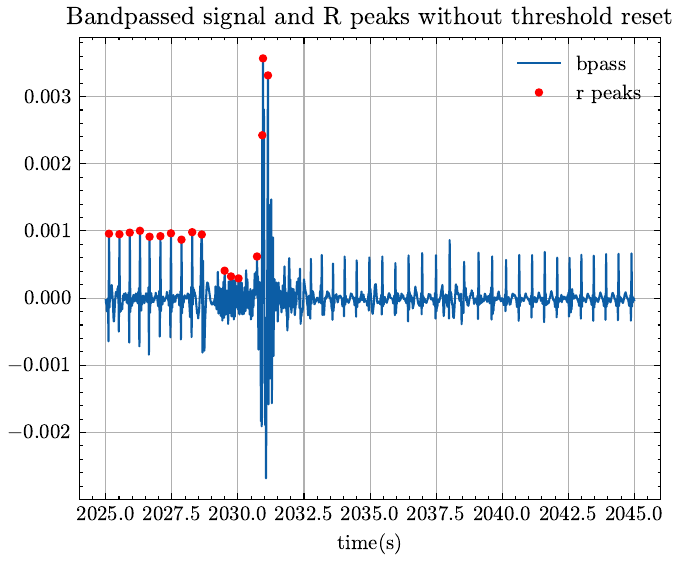}
		\caption{}
		\label{w/o_thresld_reset}
	\end{subfigure}
	\hfill
	\begin{subfigure}[b]{0.45\textwidth}
		\includegraphics[width=\textwidth]{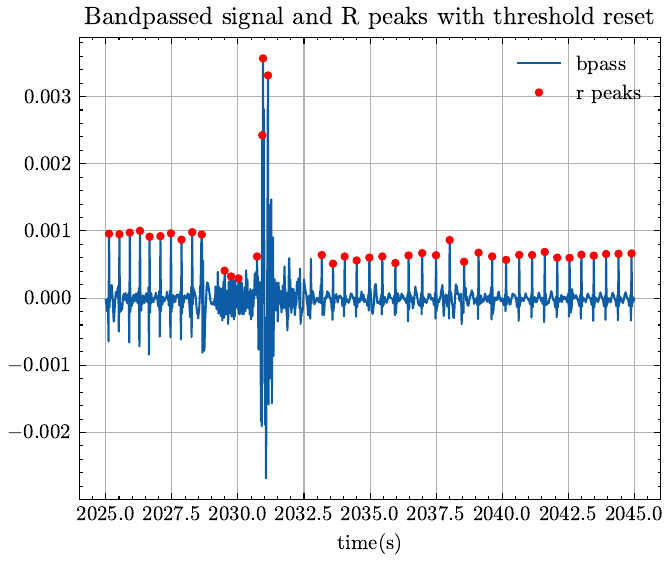}
		\caption{}
		\label{w_thresld_reset}
	\end{subfigure}
	\caption{An example of a bandpass filtered ECG signal and detected R peaks with and without thresholds reset. \textbf{(a)} Detected R peaks without thresholds reset. A high spike caused many following normal peaks missed. \textbf{(b)} After thresholds reset, these missing peaks are detected again.}
	\label{thresholds_reset_comparsion}
\end{figure}

\cref{w/o_thresld_reset} illustrates an bandpass filtered ECG segment containing a high-amplitude artefact. Following this spike, the standard algorithm fails to detect subsequent R-peaks. However, as shown in \cref{w_thresld_reset}, after applying the threshold reset mechanism, the previously missed R-peaks are successfully detected.

\textbf{Ignore zero ECG segments:} During prolonged recordings, electrode disconnection may occur, resulting in near-zero signal levels. While these segments are not absolute zero, they typically consist of very low-amplitude random noise. In this algorithm, if no RR intervals are detected for an extended period, the thresholds are reset, leading to inefficient and unnecessary processing of meaningless noise. To mitigate this issue, two measures are implemented. First, during peak detection, any potential peaks (from the moving integral signal) with amplitudes lower than $10^{-12}$ are ignored, effectively skipping most prolonged zero ECG segments. Second, for shorter zero segments (where noise exhibits slightly higher amplitude), the search-back procedure is halted if no RR intervals are detected for more than 1.4 seconds. However, the threshold reset mechanism remains active, collaborating with normal RR limits to ensure that detection resumes promptly once the signal returns to normal. These enhancements improve both the efficiency and accuracy of the algorithm when handling zero-signal segments.

\begin{figure}[htbp]
	\centering
	\begin{subfigure}[b]{0.45\textwidth}
		\includegraphics[width=\textwidth]{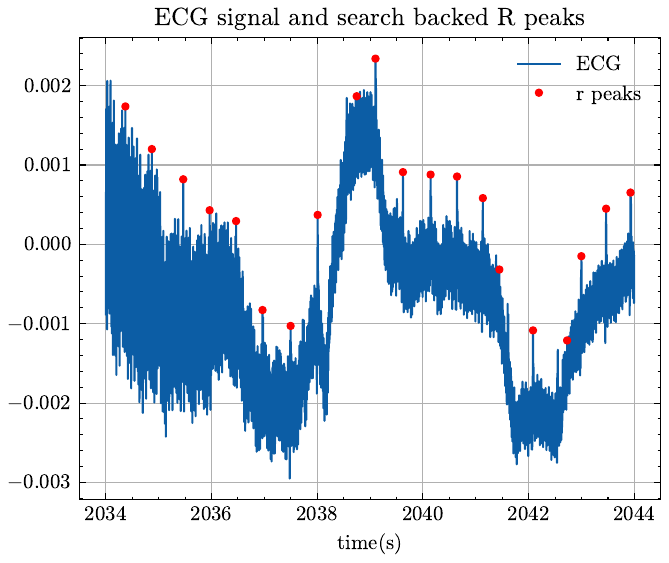}
		\caption{}
		\label{ECG_searchback}
	\end{subfigure}
	\hfill
	\begin{subfigure}[b]{0.45\textwidth}
		\includegraphics[width=\textwidth]{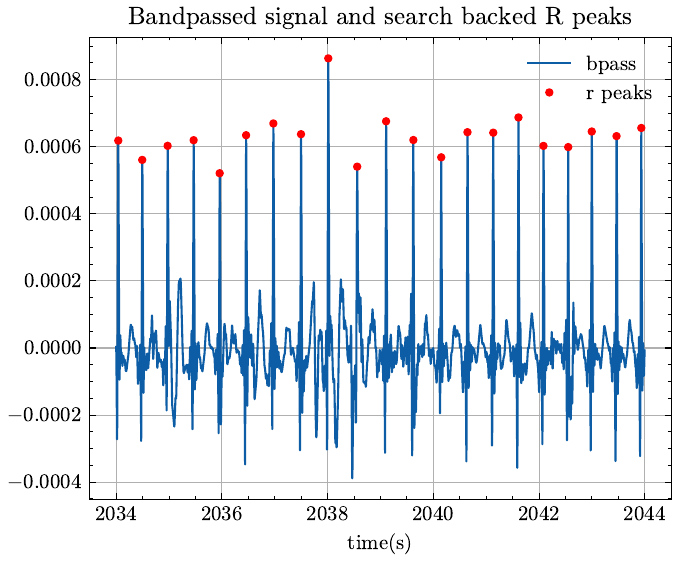}
		\caption{}
		\label{bpass_searchback}
	\end{subfigure}
	\caption{Detected R peaks search back from \textbf{(a)} raw ECG signal  and \textbf{(b)} bandpass filtered ECG signal. Due to noise and artefact, R peaks in the raw ECG signal almost cannot be identified, whereas in the R peaks from the bandpass filtered signal can be accurately recognized.}
	\label{ECG_bpass_searchback}
\end{figure}

\textbf{Bandpass filtered signal search back:} In the standard Pan-Tompkins algorithm, detected R-peaks are calibrated using the raw ECG signal. However, when the ECG signal is contaminated with noise or artefacts, R-peaks may be inaccurately located or even entirely wrong. In the proposed algorithm, the bandpass-filtered ECG signal is used instead of the raw signal for search-back correction. \cref{ECG_bpass_searchback} illustrates the final detected R-peaks using both the raw and bandpass-filtered ECG signals. Due to the presence of noise and artefacts, R-peaks are difficult to distinguish in the raw ECG signal. In contrast, after bandpass filtering, noise and artefacts are significantly reduced, allowing for more precise and reliable R-peak detection.

\subsubsection{Improvements on post-processing}
Although the improved Pan-Tompkins algorithm has significantly enhanced the quality and reliability of detected RR intervals, irregular RR intervals may still occur due to residual artefacts that cannot be completely removed. The standard Pan-Tompkins algorithm does not include a post-processing stage. However, to further improve the quality of detected RR intervals and preserve the temporal structure of RR sequence as accurate as possible, an artefact correction procedure was incorporated into the enhanced algorithm to minimize distortions caused by artefacts, making the detected RR intervals more representative of the true physiological signals.

To achieve this, irregular RR intervals are categorized into four groups: a). extremely short intervals (RR $\le$ 0.2\,s); b). short intervals (0.2\,s < RR $\le$ 2\,s); c). long intervals (2\,s < RR $\le$ 10\,s); d). extremely long intervals (RR > 10\,s). 

For extremely short intervals (RR$\le$0.2\,s), which are physiologically impossible to occur a QRS complex within such a short timeframe. Rather than removing them, they were replaced by a moving-average estimate. As direct removal of such intervals would artificially inflate adjacent intervals and distort the local heart rate dynamics, this approach can preserve temporal continuity in the RR sequence. The replacement therefore aims to maintain the overall trend of the signal rather than the exact timing of individual beats.

For short intervals (0.2\,s<RR$\le$2\,s), irregular values are replaced with the corresponding moving average value. Both adults and neonates may occasionally miss a beat. However, it is rare to miss more than one beat consecutively. Therefore, any RR interval within this range and exceeding 2.05 times the mean RR interval is classified as irregular and corrected accordingly.

For long intervals (2\,s<RR$\le$10\,s), replacing them with a moving average value is not suitable, as such cases likely indicate multiple missing beats. Instead, missing R-peaks are reconstructed using potential R-peaks extracted from the bandpass-filtered ECG signal. This approach is more reliable since potential peaks within the bandpass-filtered signal are identified as local maxima based on a predefined search window. However, due to residual artefacts and noise, not all potential peaks perfectly align with the actual QRS complexes. To ensure that only valid RR intervals are retained, any reconstructed RR interval shorter than 0.6 times the mean RR interval (the one-hour global mean) is discarded, ensuring that the inserted RR intervals remain within a physiologically reasonable range.

\begin{figure}[htbp]
	\centering
	\begin{subfigure}[b]{0.45\textwidth}
		\includegraphics[width=\textwidth]{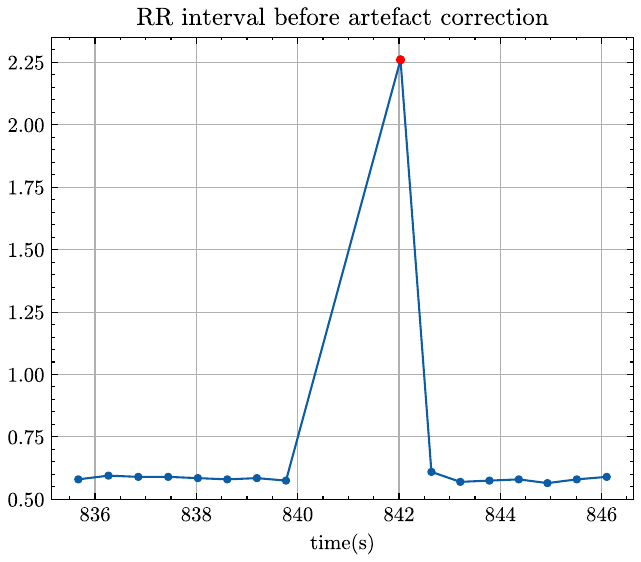}
		\caption{}
		\label{insert_peak_before_rr}
	\end{subfigure}
	\hfill
	\begin{subfigure}[b]{0.45\textwidth}
		\includegraphics[width=\textwidth]{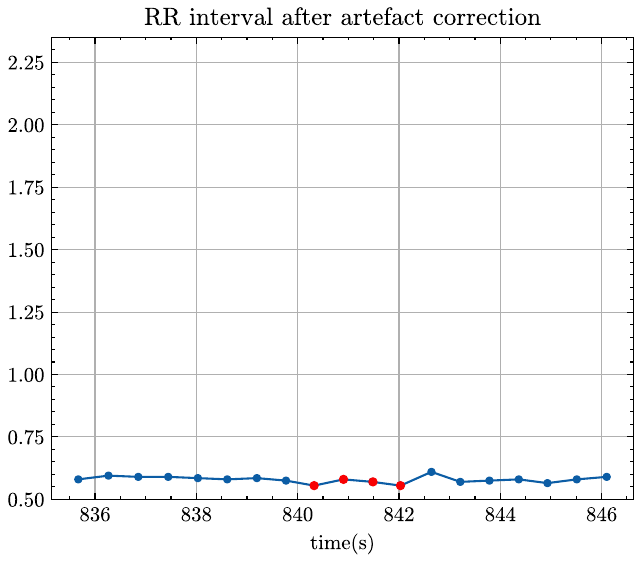}
		\caption{}
		\label{insert_peak_after_rr}
	\end{subfigure}
	\caption{An example of a long RR interval was replaced by a segment of potential R peaks. \textbf{(a)} a 2.25\,s long RR interval before artefact correction. \textbf{(b)} after artefact correction the long interval was replaced by a segment of short intervals (marked as red) from potential peaks.}
	\label{insert_peaks_rr}
\end{figure}

\cref{insert_peaks_rr} illustrates the correction of a 2.25\,s irregular RR interval, which has been replaced with several of short intervals from a segment of potential R-peaks. After artefact correction, the characteristics of the corrected segment closely resemble those of normal RR intervals, demonstrating minimal deviation.

\begin{figure}[htbp]
	\centering
	\begin{subfigure}[b]{0.45\textwidth}
		\includegraphics[width=\textwidth]{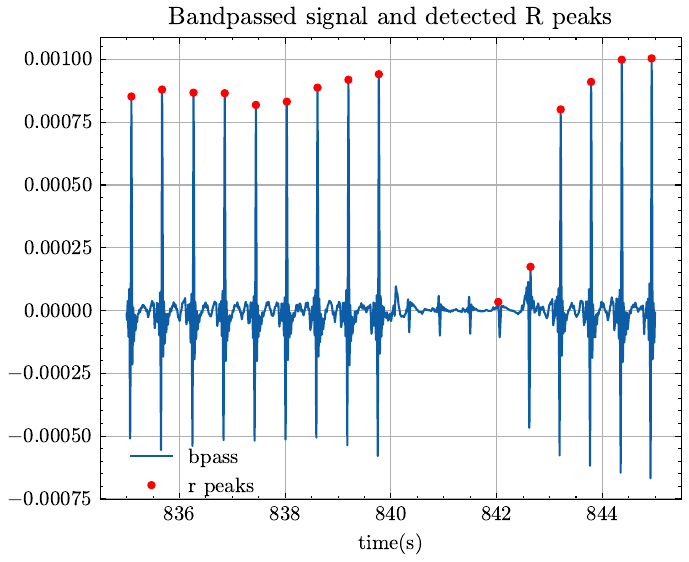}
		\caption{}
		\label{insert_peak_before_bpass}
	\end{subfigure}
	\hfill
	\begin{subfigure}[b]{0.45\textwidth}
		\includegraphics[width=\textwidth]{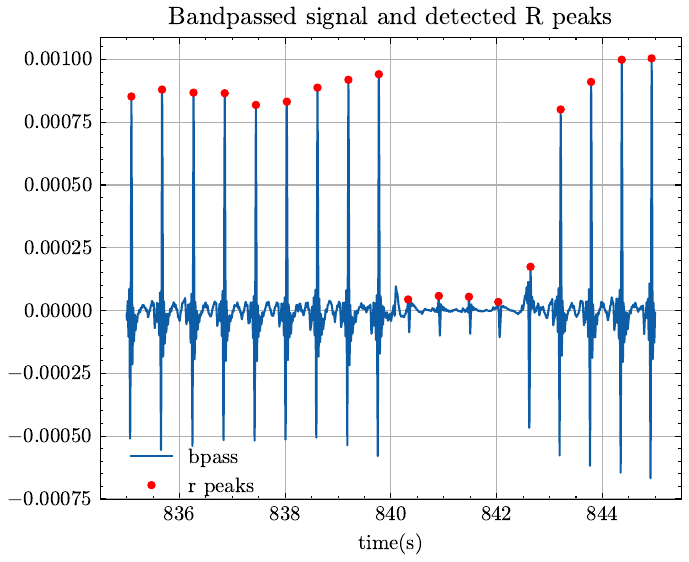}
		\caption{}
		\label{insert_peak_after_bpass}
	\end{subfigure}
	\caption{artefact correction of a long interval from bandpass filtered	ECG signal. \textbf{(a)} detected R peaks with an irregular interval caused by artefact before artefact correction. \textbf{(b)} The irregular interval was corrected by potential peaks.}
	\label{insert_peaks_bpass}
\end{figure}

\cref{insert_peaks_bpass} highlights the positions of the inserted peaks within this segment, derived from the bandpass-filtered ECG signal. Despite the presence of artefacts, the algorithm accurately identifies and aligns the reconstructed R-peaks with the true QRS complexes.

\begin{figure}[htbp]
	\centering
	\begin{subfigure}[b]{0.45\textwidth}
		\includegraphics[width=\textwidth]{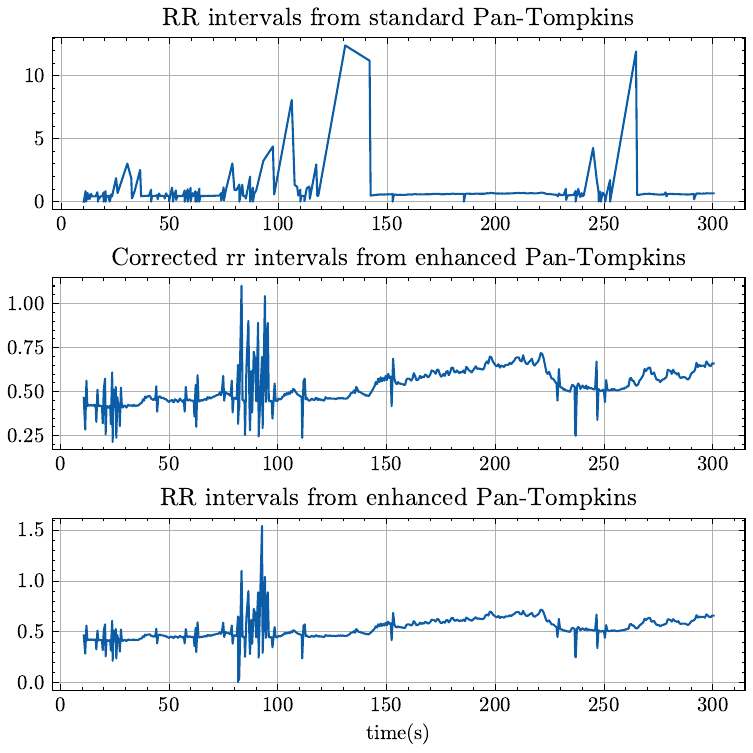}
		\caption{}
		\label{rr_comparison_example_1}
	\end{subfigure}
	\hfill
	\begin{subfigure}[b]{0.45\textwidth}
		\includegraphics[width=\textwidth]{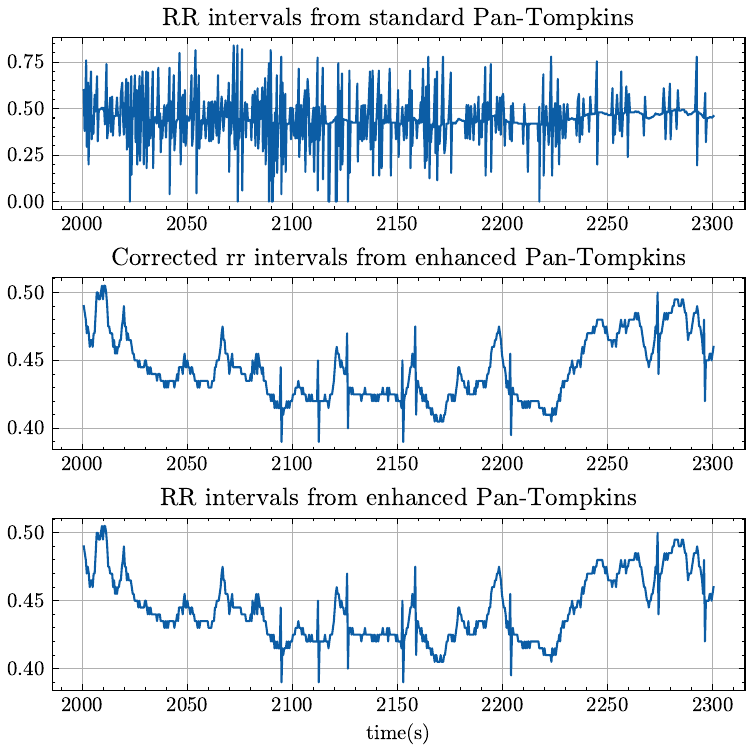}
		\caption{}
		\label{rr_comparison_example_2}
	\end{subfigure}
	\caption{Examples of extracted RR intervals comparison from standard and enhanced Pan-Tompkins as well as the artefact corrected. (more detailed examples can be seen in \protect\hyperlink{https://github.com/syu-kylin/enhanced-Pan-Tompkin}{Github})}
	\label{rr_examples}
\end{figure}

For RR intervals exceeding 10 seconds, no modifications are applied, as any artificial adjustments during such long intervals may introduce further inaccuracies. Instead, these intervals are flagged for exclusion in subsequent processing to prevent potential divergence. \cref{rr_examples} demonstrates two examples of RR intervals extracted from standard and enhanced Pan-Tompkins. The enhanced Pan-Tompkins successfully recovers long RR segments that the standard method fails to detect and markedly suppresses the noise.  

\subsection{HR Signal Preprocessing}
Before being fed into the model, the HR signal undergoes preprocessing, including resampling, segmentation, noise removal, and normalization. Despite being processed by the enhanced Pan-Tompkins algorithm, some one-hour HR epochs remain noisy due to poor ECG signal quality. To address this, each one-hour RR interval sequence is split into some long segments based on the presence of irregular RR intervals (e.g., those exceeding 4 seconds). For instance, if an one-hour epoch contains an RR interval longer than 4 seconds, it is divided into two long segments, effectively excluding that interval. Resampling is performed using two-step interpolation. First, each long segment undergoes linear interpolation to 256\,Hz to preserve all relevant information as much as possible. This is followed by cubic spline interpolation at a final sampling frequency of 4\,Hz \cite{vestOpenSourceBenchmarked2018}, as frequencies up to 2\,Hz are considered the most relevant for HRV analysis \cite{gouldingHeartRateVariability2015}.

Each one-hour epoch (might include several long segments) is then segmented again into overlapping 5-minute windows with an 80\% overlap to increase the number of training samples. A 5-min window is widely recommended for short-term HRV analysis \cite{shaffer2017overview, voss2015short, adam2023heart}. These windows serve as individual input samples for the model, inheriting the label of their corresponding one-hour epoch. Noise removal is applied at the sample level to maximize data utilization. Any sample with a standard deviation exceeding a specific value (e.g., 0.12 -- variable value was empirically chosen to balance the signal quality and data retention) is removed. Additionally, one-hour epochs containing fewer than 10 valid windows are removed. Finally, all window samples are normalized using min-max normalization. To mitigate the impact of outliers, the 5\th{} and 95\th{} percentiles of the dataset are used as the minimum and maximum values, respectively. After preprocessing, the ANSeR1 and ANSeR2 strong label groups yielded 215 and 257 one-hour epochs of RR intervals, respectively. Meanwhile, the weak label groups provided 1,316 one-hour epochs from ANSeR2.

\subsection{HRVConformer}
\begin{figure}[htbp]
\centering
\includegraphics[width=10 cm]{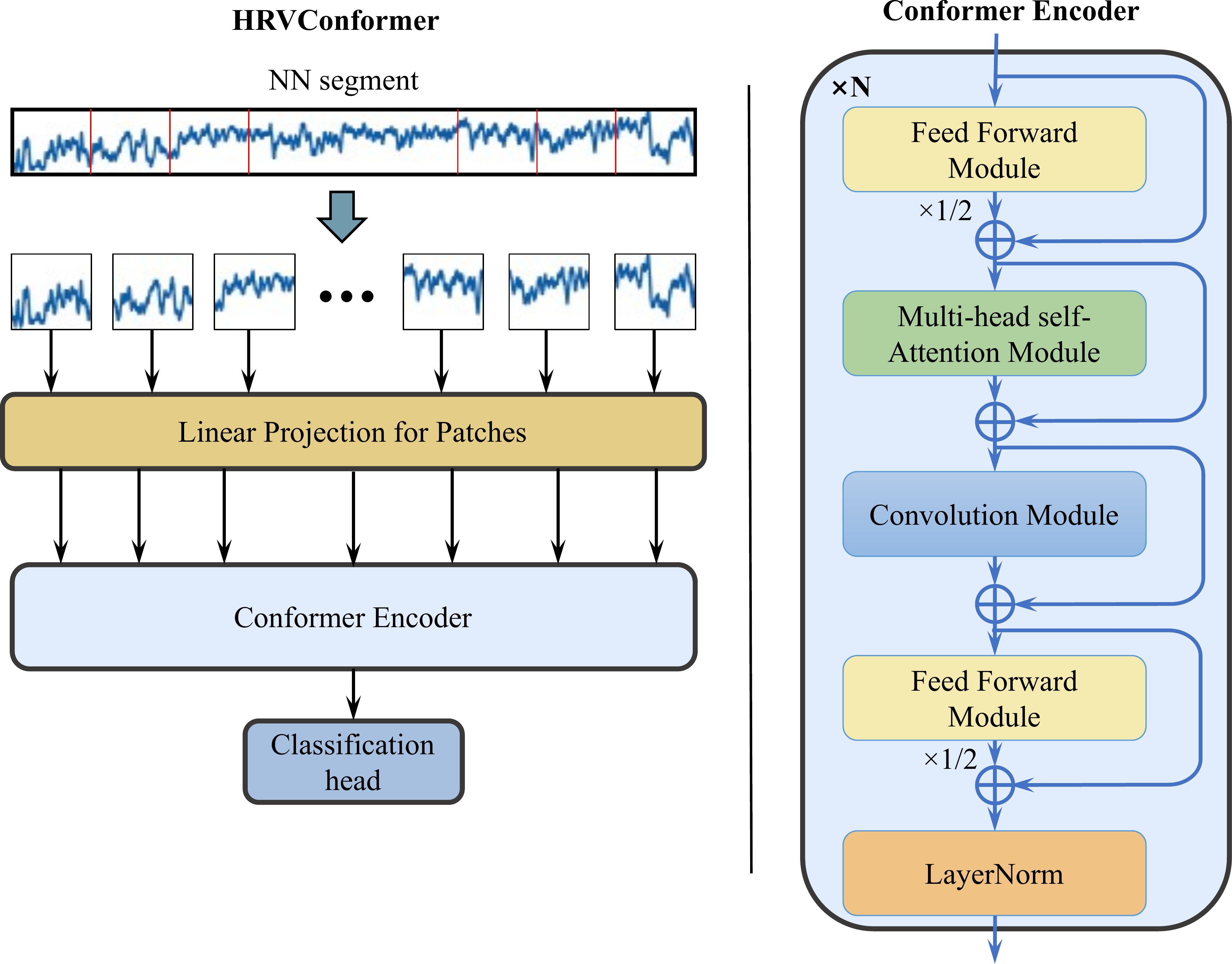}
\caption{Overview of HRVConformer model architecture. A 5-minute NN-intervals sample is split into a fixed length of patches for HRV feature representation. After linearly embedding, these vectors are fed into a Conformer encoder. A classification head is added to perform the HIE classification task.}
\label{HRVConformer}
\end{figure}

An overview of the model architecture is depicted in \cref{HRVConformer}. Inspired by the Vision Transformer \cite{dosovitskiyImageWorth16x162021}, the 1D 5-minute heart rate (RR intervals) sample is divided into fixed-length patches with a shape of $(L \times p_s)$. Instead of using a standard Transformer encoder, the Conformer encoder \cite{gulatiConformerConvolutionaugmentedTransformer2020} was adopted, which better captures both local and global dependencies.

A linear layer project these patches into a sequence of embedding with a shape of $(L \times d_{\mathrm{model}})$. This embedding sequence is then processed by N layers of identically stacked Conformer block, and the final output with the same shape as the input embedding is fed into a classification head for HIE injury detection.

\begin{figure}[htbp]
\centering
\includegraphics[width=15 cm]{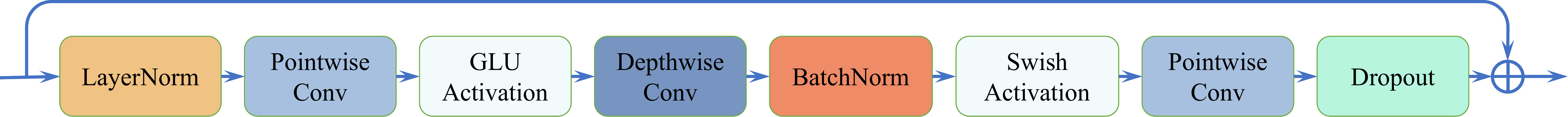}
\caption{Convolution module of Conformer block.}
\label{convolution module}
\end{figure}

Unlike the standard Transformer encoder, the Conformer encoder integrates an additional convolutional module and two half-step feedforward modules in place of the single feedforward module. The convolutional module, illustrated in \cref{convolution module}, begins with a pointwise convolution that expands the channel dimension by a factor of 2, followed by a nonlinear Gated Linear Unit (GLU), which enhances feature selection through a gating mechanism. A 1D depthwise convolution is then applied to refine local features, followed by batch normalization to stabilize training. Finally, the SiLU (Swish) activation function is used, followed by another pointwise convolution to complete this transformation.

\begin{figure}[htbp]
\centering
\includegraphics[width=14 cm]{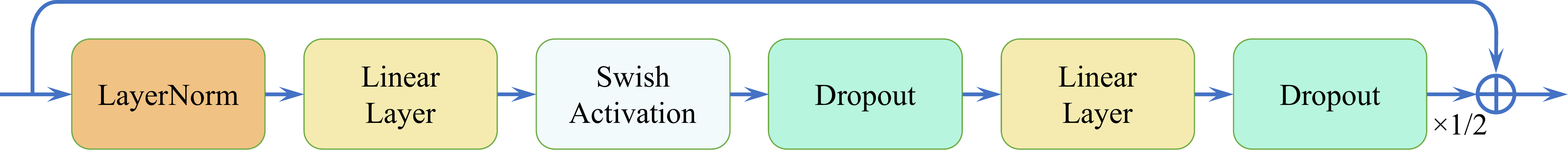}
\caption{Feedforward module of Conformer block.}
\label{feedforward module}
\end{figure}

Two feedforward modules are positioned before the attention module and after the convolution module, each applying a residual connection with a scaling factor of 0.5. The first linear layer expands the feature dimension by a factor of 4, while the second linear layer projects it back to the original $d_{\mathrm{model}}$ dimension. The SiLU (Swish) activation function is used with a pre-normalization (pre-norm) strategy to enhance stability. Dropout is applied after each linear layer to regularize the model and prevent overfitting. The feedforward structure is illustrated in \cref{feedforward module}.

\begin{figure}[htbp]
\centering
\includegraphics[width=10 cm]{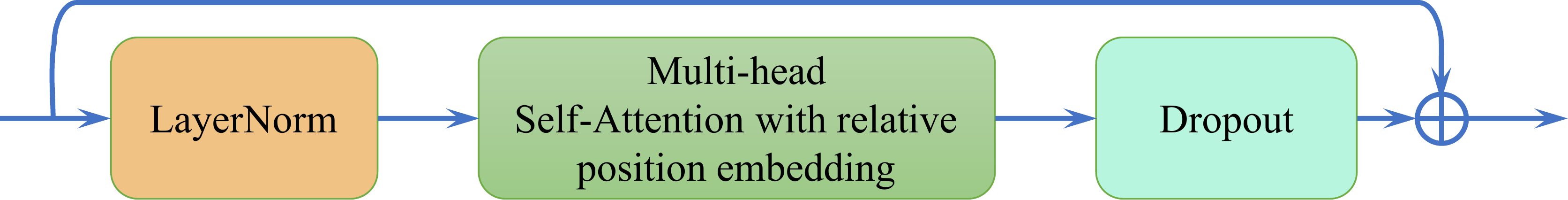}
\caption{Multi-head self-attention with relative position embedding.}
\label{MHSA}
\end{figure}

In the standard Conformer architecture, multi-head self-attention is combined with a relative positional embedding to better generalize across varying utterance lengths \cite{daiTransformerXLAttentiveLanguage2019}. Here, this technique is retained, but aims to help the attention module perceive the relative position between a clinical event and the surrounding signal. The multi-head self-attention module is illustrated in \cref{MHSA}. The residual connection with pre-norm is used to stabilize and speed up training. Dropout is followed by attention module to regularize the model.

\subsection{Post-processing}
For each one-hour epoch, the model generates multiple sample-level predictions. To determine the final classification for the epoch, a majority voting approach is applied. The final prediction is assigned based on the most frequent segment predictions, effectively mimicking the decision-making process of expert annotation. In order to obtain the AUC on the one-hour epoch level, all samples' prediction probabilities from this one-hour epoch are averaged as the prediction probability of this epoch.

\section{Experiments \& Results}
\subsection{HR data from improved Pan-Tompkins}
To evaluate the performance of the improved Pan-Tompkins algorithm, a quantitative comparison based on two key aspects is conducted: (1) the amount of usable HR data extracted from the ECG signal and (2) the quality of features derived from the HR data.

Following Pan-Tompkins processing, some RR intervals exhibit excessive noise, rendering them unsuitable for further analysis. To address this, an irregular RR exclusion and noise removal procedure were applied to filter out unreliable HR data, as described in the preprocessing section. Each one-hour epoch of RR intervals was segmented using a fixed-length window with a specified overlap, preserving as much HR data as possible while ensuring compatibility for feature extraction and model input. After segmentation, epochs with a standard deviation exceeding 0.12 were removed, and those containing fewer than 10 windows were discarded. \cref{tab:HRV data pan-tompkin comparison} presents a comparison of data availability between the standard and improved Pan-Tompkins algorithms after noise removal. The ANSeR1 and ANSeR2 strong label groups originally contained 216 and 259 one-hour epochs, respectively. However, after applying the noise removal procedure, the standard Pan-Tompkins algorithm retained only 127 and 159 one-hour epochs for ANSeR1 and ANSeR2, respectively. In contrast, the improved Pan-Tompkins algorithm preserved nearly all the data, discarding only 1 and 8 one-hour epochs, demonstrating its effectiveness in maintaining HR data availability.

\begin{table}[htbp]
 \caption{HR data availability from standard and enhanced Pan-Tompkins}
  \centering
  \begin{tabular}{lcc}
    \toprule
             & ANSeR1  & ANSeR2  \\
    \midrule
    Total    & 216h    & 259h     \\
    standard Pan-Tompkins    & 127h  & 159h      \\
    enhanced Pan-Tompkins    & 215h  & 251h  \\
    \bottomrule
  \end{tabular}
  \label{tab:HRV data pan-tompkin comparison}
\end{table}

\begin{figure}[htbp]
\centering
\includegraphics[width=10 cm]{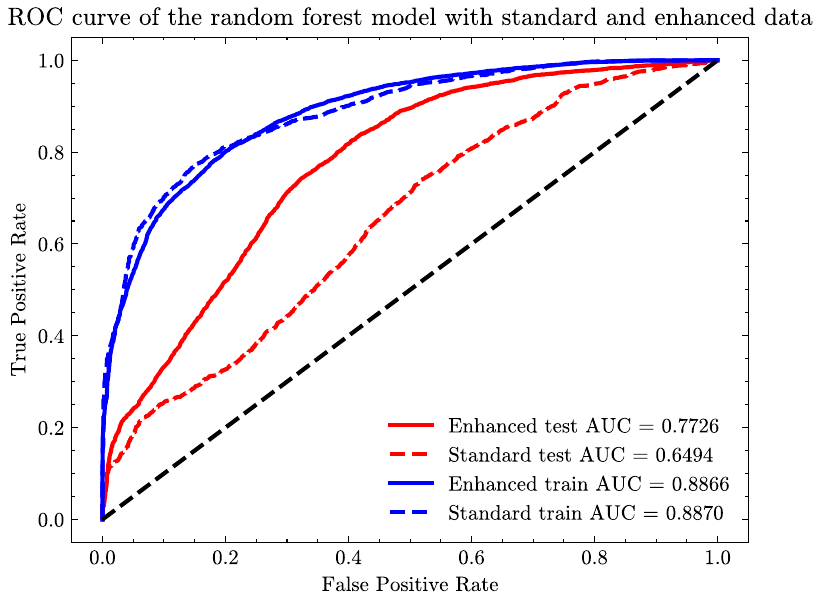}
\caption{Train and test AUC of random forest model with HR data from standard and improved version of Pan-Tompkins. Two models are fully trained to a similar train AUC, but a much lower test AUC was assessed for model trained from the standard version of Pan-Tompkins.}
\label{randomforest AUC}
\end{figure}

\begin{figure}[htbp]
\centering
\includegraphics[width=10 cm]{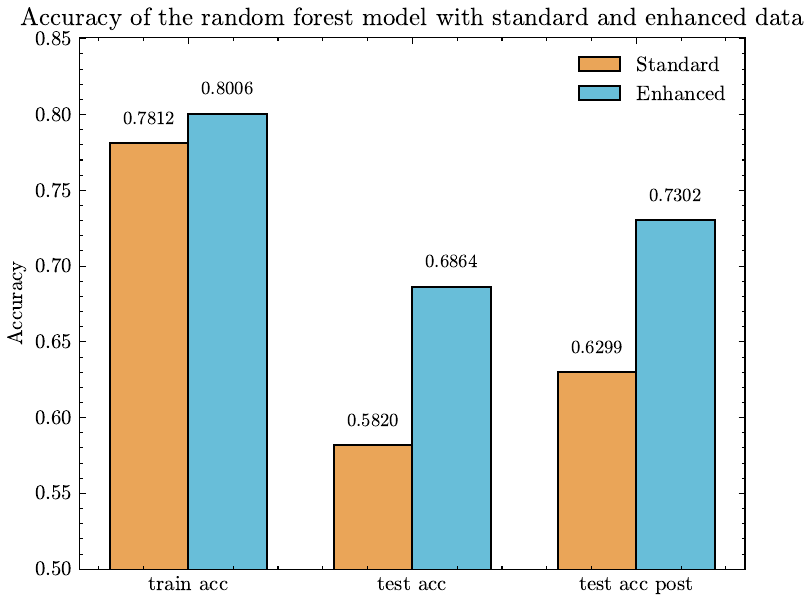}
\caption{Train and test accuracy of random forest model with HRV data from standard and enhanced Pan-Tompkins. The test acc post represents the test accuracy after the post-processing of majority vote (on the one-hour epoch level). The performance of model trained from enhanced Pan-Tompkins significantly surpass the model from standard one on all metrics.}
\label{randomforest accuracy}
\end{figure}

To assess the quality of HR data obtained from both version of the Pan-Tompkins algorithms, sixteen features from the time and frequency domains (common features used for the analysis of neonatal seizure, HIE classification \cite{rezaeiAssessingEffectivenessHeart2024a}, and preterm baby health outcome assessment \cite{semenovaPredictionShorttermHealth2019}) were extracted to represent the heart rate variability and used as input for a random forest model. The model was trained on the ANSeR2 strong label group dataset and evaluated on the ANSeR1 dataset. \cref{randomforest AUC} and \cref{randomforest accuracy} illustrate the performance of the model in terms of AUC and accuracy on both the training and test sets. The results clearly show that the random forest model trained on HR data extracted using the improved Pan-Tompkins algorithm significantly outperforms the one using the standard version. The enhanced algorithm yields higher AUC and accuracy even on a bigger test set, despite being trained to a similar training AUC, demonstrating its effectiveness in producing higher-quality HRV features.

\subsection{Experiments setup}
The development set consisted of the ANSeR2 weak and strong label group datasets. From this, 314 (20\%) one-hour epochs were randomly selected for validation, while the remaining 1,259 epochs were used for training. This large validation set provided a reliable performance estimation during training. The independent ANSeR1 strong label group, comprising 215 epochs, was reserved for testing. The model was implemented with PyTorch and trained on a SLURM system with a NVIDIA L40S GPU. A cosine warmup learning rate scheduler was used, with 50-epoch warmup period and a minimum learning rate of 1e-6. The AdamW optimizer was employed, which decouples weight decay from the learning rate, with a learning rate of $6\times 10^{-5}$ and $\beta1$, $\beta2$ values of 0.85 and 0.998, respectively. The model was trained for 1,800 epochs with a batch size of 1,024 (cost approximate 1.5h). Weight decay (0.1), dropout (0.3) and label smoothing (0.2) were applied to regularize the model, and an early stopping mechanism was implemented. The best model was selected based on the highest moving average validation AUC.

\subsection{Baseline comparison}
As this is the first time to create such a large HRV dataset, no previous studies use this dataset for HIE classification with heart rate signal. To ensure a fair and comprehensive evaluation of the proposed architecture, several alternative models are implemented for comparison. The first baseline is a Transformer-based model, denoted as HRVTransformer, in which each Conformer block is replaced with a standard Transformer block while keeping the remaining configurations identical to the HRVConformer. This design isolates the effect of incorporating convolutional components within the Transformer structure and enables direct comparison of their contributions to HIE classification.

Two purely convolutional architectures were also evaluated. The first, FCN19 introduced in \cite{yu2023neonatal}, is a fully convolutional network comprising six identical feature extraction blocks followed by a classification block. Each feature extraction block contains three 1D convolutional layers with ReLU activations, followed by a batch normalization, and average pooling, all using a fixed feature dimension of 64. The classification block replaces fully connected layers with a convolutional projection that maps the embedding size (64) to the number of output classes.

The second convolutional baseline is a ResNet-based model, HRVRes50, adapted from the original ResNet-50 architecture to 1D inputs. It employs bottleneck blocks to process each 5-minute HR segment as a one-dimensional image, while retaining the channel dimensions of the original design. This adaptation allows the model to exploit the hierarchical feature extraction strengths of ResNet for time-series HR signal classification. These baselines provide complementary perspectives for benchmarking HRVConformer and allow a rigorous evaluation of its effectiveness. Different model configurations are demonstrated in \cref{tab:model_configurations}. The configuration of HRVConformer is based on a hyperparameter tuning on the validation set.

\begin{table}[htbp]
	\caption{Different model configurations.}
	\centering
	\begin{tabular}{ll}
		\toprule
		models & configuration \\
		\midrule
		HRVConformer & 
		\makecell[l]{
			patch size: 25s \\
			embed dim: 144  \\
			number of layers: 3 \\
			number of attention heads: 8 \\
			depthwise conv kernel size: 11 \\
			position embedding: relative \\
			classifier head: FCN \\
			fcn head conv kernel size: 11 \\
		} \\
		\midrule
		
		HRVTransformer &
		\makecell[l]{
			patch size: 25s \\
			embedding dim: 144  \\
			number of layers: 3 \\
			number of attention heads: 8 \\
			classifier head: FCN \\
			fcn head conv kernel size: 11 \\
			position embedding: none \\
		} \\
		\midrule
		
		HRVRes50 & 
		\makecell[l]{
			number of blocks: 5 \\
			block type: bottleneck \\
			block embed dim: [64, 128, 256, 512]
		} \\
		\midrule
		
		FCN19 & 
		\makecell[l]{
			number of layers: 19 \\
			embedding dimension: 64
		} \\
		\bottomrule
	\end{tabular}
	\label{tab:model_configurations}
\end{table}

\cref{tab:baseline accuracy comparison} summarizes the validation and test performance of all models, averaged over ten randomized experiments (five with different validation splits and five with different random seeds). For the HRVTransformer model, positional encoding was disabled (as discussed later). Across all experiments, the HRVConformer consistently achieved the best performance on both validation and test sets. On the test set, the HRVTransformer closely followed, attaining accuracy comparable to the HRVConformer but with slightly lower AUC. Both Transformer-based models (HRVConformer and HRVTransformer) outperformed the convolutional baselines (FCN19 and HRVRes50). Interestingly, despite having substantially more parameters, the HRVRes50 underperformed compared with the FCN19. In validation, the HRVTransformer was outperformed by both convolutional models, highlighting that the HRVConformer likely inherits the complementary strengths of convolutional and Transformer architectures --- efficient learning and strong generalization.

\begin{figure}[htbp]
	\centering
	\includegraphics[width=10 cm]{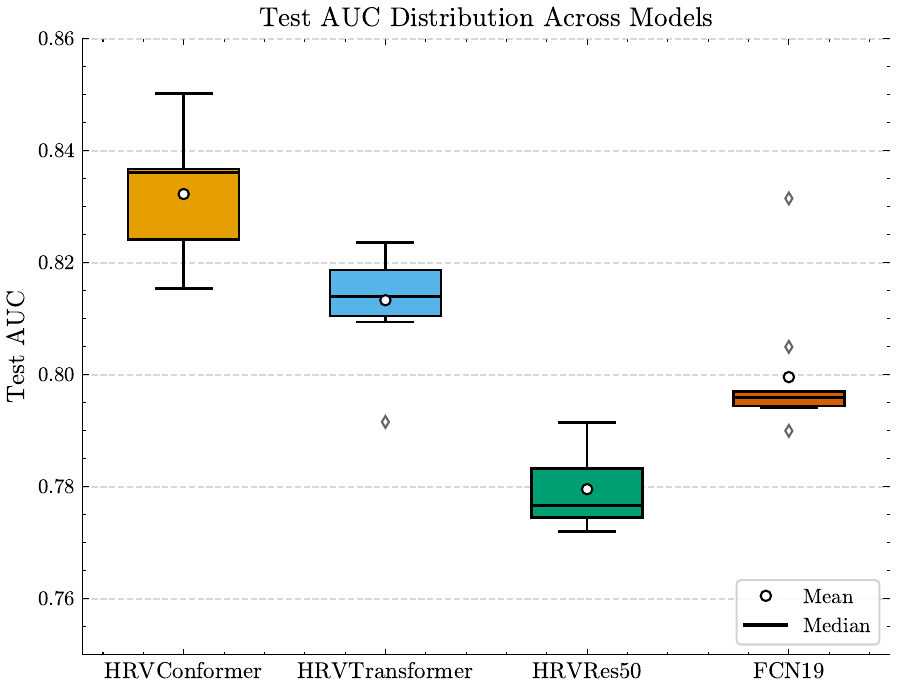}
	\caption{Different models test ROC-AUC distribution (epoch level). Each model run with 10 times random experiments.}
	\label{model test AUC distribution}
\end{figure}

The distribution of the test AUCs (at an epoch level) across the ten runs is shown in \cref{model test AUC distribution}. The HRVConformer achieved an median AUC of 83\%, significantly surpassing all baselines. Although the HRVTransformer underperformed the HRVConformer, it exhibited a narrower spread across runs. The FCN19 showed the most stable distribution overall, though with occasional high outliers, the best of which approached the mean performance of the HRVConformer, suggesting latent potential.

\begin{figure}[htbp]
\centering
\includegraphics[width=10 cm]{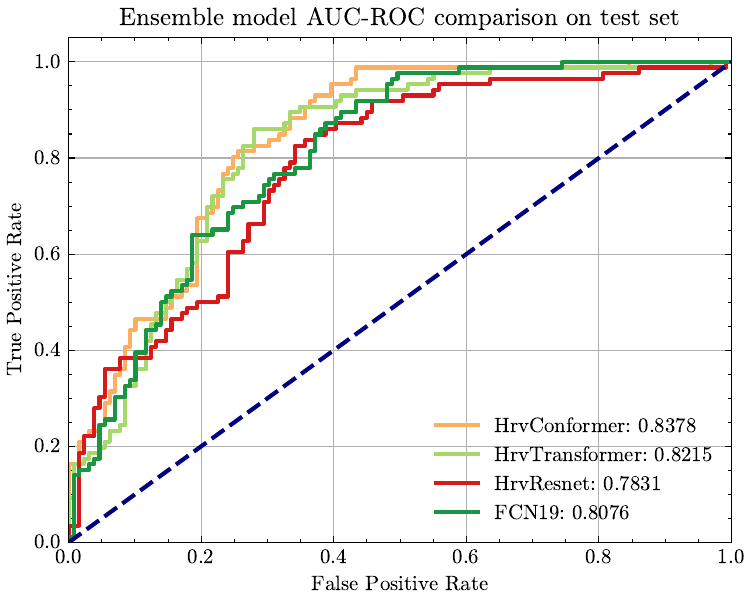}
\caption{Different ensemble models test ROC-AUC comparison on epoch level (from 10 random experiments for each model).}
\label{model AUC comparison}
\end{figure}

\cref{model AUC comparison} presents the ROC curves of the 10 ensembled models on the epoch level. The probability of each one-hour epoch is averaged over all 10 model predictions. The HRVConformer achieved the highest test AUC, followed by the HRVTransformer and the FCN19, while the HRVRes50 performed the worst.

\begin{table}[htbp]
 \caption{Different model validation and test performance comparison (epoch level).}
  \centering
  \begin{tabular}{lcccc}
    \toprule
                    &        test ACC     &      test AUC       &     val ACC         & val AUC \\
    \midrule
    HRVConformer    & $\mathbf{0.7456_{\pm0.016}}$ & $\mathbf{0.8323_{\pm0.010}}$ & $\mathbf{0.8787_{\pm0.014}}$ & $\mathbf{0.9524_{\pm0.007}}$ \\
    HRVTransformer  & $0.7428_{\pm0.023}$ & $0.8133_{\pm0.009}$ & $0.7908_{\pm0.031}$  & $0.8868_{\pm0.019}$ \\
    HRVRes50        & $0.7033_{\pm0.015}$ & $0.7796_{\pm0.007}$ & $0.8315_{\pm0.018}$ & $0.9150_{\pm0.008}$\\
    FCN19           & $0.7144_{\pm0.021}$ & $0.7996_{\pm0.011}$ & $0.8691_{\pm0.007}$ & $0.9463_{\pm0.007}$\\ 
    \bottomrule
  \end{tabular}
  \label{tab:baseline accuracy comparison}
\end{table}

\subsection{Ablation study}
To assess the contribution of individual components within the proposed model architecture, a series of ablation studies was conducted. By selectively removing or modifying key elements, their respective impacts on model performance was evaluated. These experiments provide valuable insights into the critical components of the HRVConformer architecture. All ablation studies were performed using the same datasets, training procedures, and evaluation metrics described in Section 3.2 to ensure consistency and comparability.

\cref{tab:ablation HRVConformer components} presents the results of a progressive ablation study where essential modules of the Conformer block were systematically removed, effectively transitioning the model toward a standard Transformer configuration. The exclusion of the half-step feed-forward and convolutional module resulted in similar performance declines 0.75\% and 0.97\% of test AUC, respectively. This suggests that both components play significant roles in capturing local and intermediate-level features. Notably, the convolution module appears to contribute slightly more to performance, likely due to its fine processing on local features. Interestingly, the removal of positional encoding did not negatively affect model performance; on the contrary, it led to a marginal improvement in test AUC (0.19\%). A similar situation was also observed with the HRVTransformer model. This indicates that, in the context of HR signals, explicit temporal position information may be less critical or even detrimental under certain conditions.

Additional ablation experiments were conducted to assess the impact of patch size and data normalization strategies (\cref{tab:ablation patch szie} and \cref{tab:ablation of normlization}). Unlike textual or visual data, HR signals lack a natural notion of segment granularity, making patch size selection non-trivial. Smaller patches yield longer sequences, increasing computational cost and risking fragmented representations of HR dynamics, whereas larger patches shorten the sequence but may obscure local variations by averaging over heterogeneous regions. Among the tested configurations, a patch size of 25\,s achieved the best performance, yielding the highest AUC on the one-hour test set. Patch sizes of 10–20\,s produced comparable results, while very small patches (5\,s) significantly degraded performance.

For normalization, min–max scaling and z-score standardization were compared. The min–max scaling transformed each signal to the [0, 1] range using the 5\th{} and 95\th{} percentiles to mitigate outliers, while z-score normalization standardized signals using the dataset-wide mean and standard deviation. Both methods improved training stability, but min–max scaling achieved substantially higher test AUC. This advantage is likely due to its smaller distributional shift on the independent test set, which originated from a different dataset. In contrast, z-score normalization may have introduced biases tied to the training distribution, reducing generalization.

\begin{table}[htbp]
	\centering
	\caption{Ablation study of HRVConformer components}
	\begin{tabular}{lc}
		\toprule
		cases    & test AUC epoch    \\
		\midrule
		Full HRVConfomer &  $0.8323_{\pm0.010}$ \\
		\midrule
		Remove half FFN  &  $0.8248_{\pm0.012}$  \\
		Remove Conv.     &  $0.8151_{\pm0.020}$  \\
		Remove PosEmd.   &  $0.8170_{\pm0.010}$  \\
		\bottomrule
	\end{tabular}
	\label{tab:ablation HRVConformer components}
\end{table}

\begin{table}[htbp]
	\centering
	\begin{minipage}{\shorttablewidth}
		\centering
		\caption{Ablation study of patch size.}
		\begin{tabular}{lc}
			\toprule
			patch size  & test AUC epoch   		 \\
			\midrule
			5s			& $0.7735_{\pm0.012}$    \\
			10s         & $0.8091_{\pm0.015}$    \\
			15s         & $0.8168_{\pm0.016}$    \\
			20s         & $0.8134_{\pm0.008}$    \\
			25s         & \colorbox{gray!30}{$\mathbf{0.8323_{\pm0.010}}$} \\
			\bottomrule
		\end{tabular}
		\label{tab:ablation patch szie}
	\end{minipage} %
	\hfill 
	\begin{minipage}{\shorttablewidth} 
		\centering
		\caption{Ablation study of normalization method.}
		\begin{tabular}{lc}
			\toprule
			norm  & test AUC epoch   \\
			\midrule
			standard        &  $0.7915_{\pm0.010}$   \\
			min-max         &  \colorbox{gray!30}{$\mathbf{0.8323_{\pm0.010}}$} \\
			\bottomrule
		\end{tabular}
		\label{tab:ablation of normlization}
	\end{minipage}
	
\end{table}

Further ablation studies on different HRVConformer configurations was performed to assess the impact of architectural design choices (\cref{tab:model params ablation}). Increasing model depth generally improved generalization, although the effect was modest after 3 blocks. In contrast, increasing model width did not yield consistent improvements, with negligible differences in test AUC observed across configurations.

Varying the number of attention heads also had limited impact. Models with 8 and 12 heads achieved marginally higher test AUC, while the 4-head configuration performed slightly worse, likely due to reduced capacity for capturing diverse attention patterns. Overall, these results suggest that HRVConformer performance is relatively robust to changes in depth, width, and attention heads, with only minor gains achieved through architectural scaling.

Three types of classifier heads were also evaluated: a fully convolutional network (FCN), a class token classifier, and a global pooling-based classifier. The FCN head comprises two convolutional layers, followed by average pooling with a stride of 4 and a global average pooling layer. The first convolution layer projects the sequence length to the number of output classes, enabling patch-wise classification. The resulting feature map is then reduced to a single value across the embedding dimension, with predictions made over the channel dimension.

By contrast, the class token classifier relies exclusively on the learned class token, projecting from the model embedding dimension ($d_{\text{model}}$) to the class space through a fully connected layer. The global pooling classifier aggregates sequence information using global average pooling across patches, followed by a fully connected layer. Experimental results show that the FCN classifier achieved the highest test AUC, although it has far fewer parameters due to the elimination of fully connected layer. The class token classifier is better than the global pooling one, which also confirms the aggregation ability of attention layers.

The impact of kernel size in the depthwise convolution layer of the convolution module on the model performance is also examined. As this layer operates along the sequence dimension, the kernel size determines the number of neighbouring patches incorporated during feature extraction. Results showed only minor differences across configurations, with kernel sizes of 5, 7 and 11 yielding slightly better generalization.

Similarly, varying the kernel size in the first convolutional layer of the FCN classifier head had negligible impact on test AUC. Unlike the Conformer module, this convolution operates over the embedding dimension rather than the sequence dimension. Kernel sizes of 7, 9 and 11 achieved marginally higher performance, but overall the effect was minimal.

\begin{table}[htbp]
	\centering
	\caption{HRVConformer ablation experiments. Experiments default use 25s patch size and min-max normalization. The default settings are marked with \colorbox{gray!30}{gray}.} 
	\label{tab:model params ablation}
	
	\begin{minipage}{\shorttablewidth}
		\centering
		\captionsetup{labelformat=empty} 
		\caption*{(a) Model depth.} 
		\begin{tabular}{lc}
			\toprule
			n blocks  & test AUC epoch       \\
			\midrule
			1         &  $0.8228_{\pm0.014}$ \\
			2         &  $0.8237_{\pm0.007}$ \\
			3         &  \colorbox{gray!30}{$0.8323_{\pm0.010}$} \\
			4         &  $\mathbf{0.8331_{\pm0.009}}$ \\
			5         &  $\mathbf{0.8364_{\pm0.006}}$ \\
			\bottomrule
		\end{tabular}
	\end{minipage} %
	\hfill
	\begin{minipage}{\shorttablewidth}
		\centering
		\captionsetup{labelformat=empty}
		\caption*{(b) Model width.} 
		\begin{tabular}{lc}
			\toprule
			$d_{model}$  & test AUC epoch \\
			\midrule
			128         &  $0.8288_{\pm0.004}$ \\
			144         &  \colorbox{gray!30}{$0.8323_{\pm0.010}$} \\
			192         &  $\mathbf{0.8373_{\pm0.012}}$ \\
			256         &  $\mathbf{0.8351_{\pm0.005}}$ \\
			\bottomrule
		\end{tabular}
	\end{minipage}  %
	\hfill
	\begin{minipage}{\shorttablewidth}
		\centering
		\captionsetup{labelformat=empty}
		\caption*{(c) attention head.}
		\begin{tabular}{lc}
			\toprule
			n heads  & test AUC epoch   \\
			\midrule
			4         &  $0.8299_{\pm0.011}$ \\
			6         &  $0.8300_{\pm0.013}$ \\
			8         &  \colorbox{gray!30}{$\mathbf{0.8323_{\pm0.010}}$} \\
			12        &  $\mathbf{0.8313_{\pm0.011}}$ \\
			\bottomrule
		\end{tabular}
	\end{minipage} %
	\hfill
	\begin{minipage}{\shorttablewidth}
		\centering
		\captionsetup{labelformat=empty}
		\caption*{(d) classifier head.}
		\begin{tabular}{lc}
			\toprule
			Classifier head  & test AUC epoch   \\
			\midrule
			FCN          & \colorbox{gray!30}{$\mathbf{0.8323_{\pm0.010}}$}  \\
			class token      & $0.8246_{\pm0.008}$  \\
			global pooling    & $0.8146_{\pm0.008}$  \\
			\bottomrule
		\end{tabular}
	\end{minipage} %
	\bigskip 
	\begin{minipage}{\shorttablewidth}
		\centering
		\captionsetup{labelformat=empty}
		\caption*{(e) Conv. module depthwise convolution kernel size.}
		\begin{tabular}{lc}
			\toprule
			& test AUC epoch   \\
			\midrule
			3         &  $0.8333_{\pm0.008}$ \\
			5         &  $\mathbf{0.8353_{\pm0.010}}$ \\
			7         &  $\mathbf{0.8337_{\pm0.006}}$ \\
			9         &  $0.8213_{\pm0.008}$ \\
			11        &  \colorbox{gray!30}{$\mathbf{0.8323_{\pm0.010}}$} \\
			\bottomrule
		\end{tabular}
	\end{minipage}%
	\hfill %
	\begin{minipage}{\shorttablewidth}
		\centering
		\captionsetup{labelformat=empty}
		\caption*{(f) FCN head convolution kernel size.}
		\begin{tabular}{lc}
			\toprule
			& test AUC epoch   \\
			\midrule
			3         &  $0.8229_{\pm0.008}$ \\
			5         &  $0.8216_{\pm0.006}$ \\
			7         &  $\mathbf{0.8305_{\pm0.002}}$ \\
			9         &  $\mathbf{0.8288_{\pm0.003}}$ \\
			11        &  \colorbox{gray!30}{$\mathbf{0.8323_{\pm0.010}}$} \\
			13        &  $0.8192_{\pm0.009}$ \\
			\bottomrule
		\end{tabular}
	\end{minipage}%
	
\end{table}

\section{Discussion}
In the standard Vision Transformer architecture, the fixed sinusoidal (sin–cos) positional embeddings are added to each patch embedding before entering the Transformer blocks. However, in the ablation study, the positional embeddings did not appear to be a crucial factor for the HRVConformer and were in some cases detrimental. Different positional encoding strategies for both HRVConformer and HRVTransformer were therefore evaluated, shown in \cref{tab:ps_embed_conformer} and \cref{tab:ps_embed_transformer}. For HRVConformer, using the relative positional embeddings or omitting them entirely produced comparable test AUCs, while for HRVTransformer, no positional embeddings yielded slightly better results than the relative embeddings. In contrast, the fixed sin–cos embeddings substantially reduced generalization performance for both models (76.99\% and 74.82\% test AUC for HRVConformer and HRVTransformer, respectively).

This finding may be explained by the nature of HR signals, where clinically relevant events can occur at any point within a segment rather than at a fixed location. Unlike the fixed embeddings, which impose absolute positional information on each patch, the relative embeddings capture only the distance between patches, making them more suitable for this domain. Furthermore, the learnable nature of relative embeddings likely provides greater flexibility than the rigid encoding imposed by sin–cos functions.

\begin{table}[htbp]
	\centering
	\begin{minipage}{\shorttablewidth}
		\centering
		\caption{HRVConformer with different position embedding.}
		\begin{tabular}{lc}
			\toprule
			position embedding           & test AUC epoch   		 \\
			\midrule
			relative 		& $\mathbf{0.8323_{\pm0.010}}$    \\
			without         & $\mathbf{0.8319_{\pm0.012}}$    \\
			fixed (sin-cos)        & $0.7699_{\pm0.028}$    \\
			\bottomrule
		\end{tabular}
		\label{tab:ps_embed_conformer}
	\end{minipage} %
	\hfill 
	\begin{minipage}{\shorttablewidth} 
		\centering
		\caption{HRVTransformer with different position embedding.}
		\begin{tabular}{lc}
			\toprule
			position embedding  & test AUC epoch   \\
			\midrule
			relative             & $\mathbf{0.8133_{\pm0.009}}$    \\
			without             & $\mathbf{0.8190_{\pm0.005}}$    \\
			fixed (sin-cos)     & $0.7482_{\pm0.029}$    \\
			\bottomrule
		\end{tabular}
		\label{tab:ps_embed_transformer}
	\end{minipage}
	
\end{table}

\begin{figure}[htbp]
	\centering
	\includegraphics[width=0.6\textwidth]{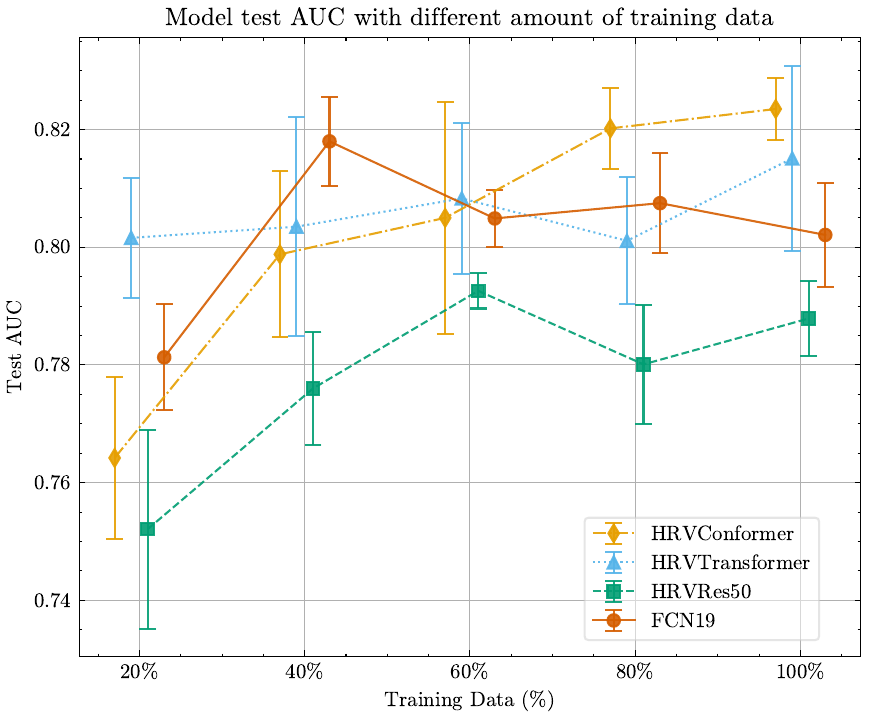}
	\caption{Model test AUC (epoch level) comparison over different amount of training data (100\% for 1259 one-hour epochs for training, validation set keeps 314 one-hour epochs). (Position encoding was removed on HRVTransformer models.)}
	\label{data_scaling_4_model}
\end{figure}

The scaling properties of HRVConformer is also examined, against the three baselines using only a fraction (20–100\%) of the training data, with a fixed validation set of 314 one-hour epochs. Each model was repeated five times with different seeds. The results are demonstrated in \cref{data_scaling_4_model}.

With 20\% of training data, HRVTransformer outperformed the other models. However, when data increased to 40\%, HRVConformer, HRVRes50, and FCN19 exhibited sharper gains, likely due to convolutional priors knowledge (inductive bias). At 60\%, the HRVConformer, HRVTransformer, and FCN19 achieved similar performance, while HRVRes50 lagged slightly but maintained steady growth. Beyond this point, HRVConformer continued to improve and surpassed all baselines, whereas FCN19 plateaued and HRVTransformer and HRVRes50 recovered from a dip at 80\%.

Overall, HRVConformer showed the strongest and most consistent scaling trend, outperforming all baselines once training data exceeded 60\%. HRVTransformer maintained a more moderate but stable improvement, only being overtaken by HRVConformer at full-scale training. FCN19 was competitive with smaller datasets but saturated with larger ones, likely due to limited capacity. HRVRes50, though weaker overall, demonstrated stable improvements with more data. These results indicate that HRVConformer likely combines the fast learning of convolutional models with the scalability of transformers.

\begin{figure}[htbp]
	\centering
	\includegraphics[width=\textwidth]{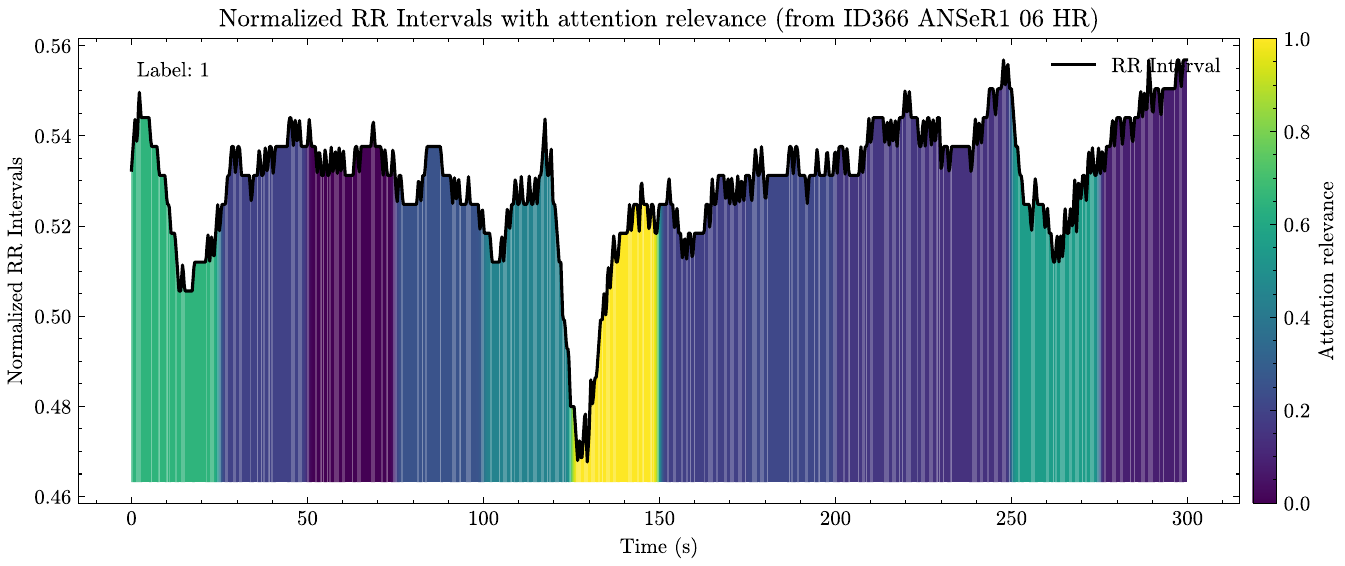}
	\caption{Example of normalized RR intervals with corresponding attention relevance. Relevance values are normalized to the range 0–1, where lighter regions indicate higher model attention and darker regions represent lower relevance.}
	\label{rr_attn_relevance}
\end{figure}

\cref{rr_attn_relevance} illustrates a normalized 5-min segment of RR intervals alongside the corresponding attention relevance. The relevance was computed using rollout attention \cite{abnar2020quantifying}, which quantifies how information propagates from the input through successive layers to the final embeddings. Notably, the intervals between 0–25\,s, 125–150\,s and 250-275\,s exhibit higher attention relevance, coinciding with more pronounced abnormalities in heart rate dynamics (further analysis might be need from the clinical perspective). This suggests that the model selectively attends to irregular HRV patterns, which may be clinically important indicators of autonomic dysfunction. By aligning attention with physiologically abnormal segments, the visualization provides interpretability and further validates the model’s capacity to capture meaningful features related to pathological changes.

\begin{figure}[htbp]
	\centering
	\begin{subfigure}[b]{0.45\textwidth}
		\includegraphics[width=\textwidth]{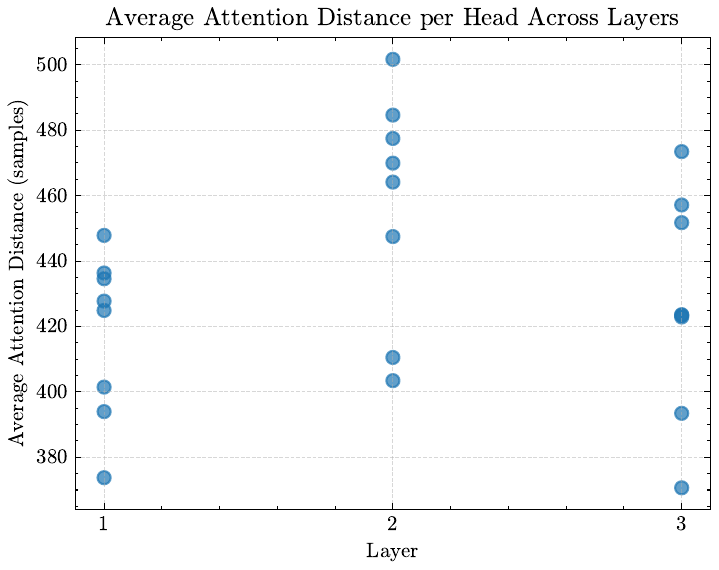}
		\caption{}
		\label{attn_dist_conformer}
	\end{subfigure}
	\hfill
	\begin{subfigure}[b]{0.45\textwidth}
		\includegraphics[width=\textwidth]{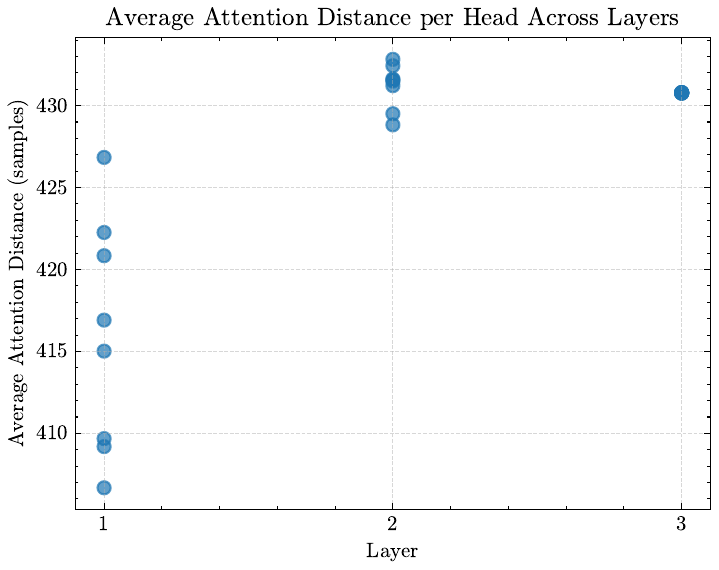}
		\caption{}
		\label{attn_dist_transformer}
	\end{subfigure}
	\caption{Average attention distance across all layers and heads, computed over the entire test set. Distances are expressed in number of input samples (5-minute for 1200 samples and 25s patch size corresponding to 100 samples). \textbf{(a)} HRVConformer shows wider and more diverse attention distances across layers. \textbf{(b)} HRVTransformer focuses on a narrower range with less variation between heads.}
	\label{attn_dist_comparison}
\end{figure}

\cref{attn_dist_comparison} compares the average attention distance of HRVConformer and HRVTransformer across all test samples, aggregated for each head and layer. The distance was computed as the absolute position difference weighted by the corresponding attention scores. HRVConformer attends over a wider range, with average distances of 370–510 samples, compared with 405–435 samples for HRVTransformer. With a patch size of 25\,s and a sampling frequency of 4\,Hz, HRVConformer focuses on 3–4 neighboring patches in the first layer, aggregates information across 4–5 patches in the intermediate layer, and attends to 3–5 patches in the final layer. In contrast, HRVTransformer consistently operates within a narrower context of about four patches, with little variation across heads. Moreover, the head-level spread of attention distance is smaller in HRVTransformer, particularly in the second and third layer where all heads converge to nearly identical distances. This redundancy suggests that HRVTransformer captures less diverse contextual information, which may explain its weaker performance relative to HRVConformer.

\begin{figure}[htbp]
	\centering
	\begin{subfigure}[b]{0.45\textwidth}
		\includegraphics[width=\textwidth]{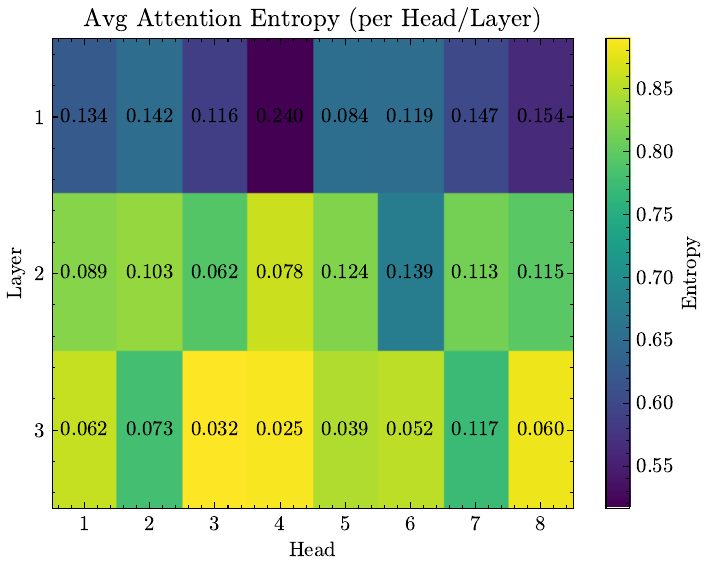}
		\caption{}
		\label{subfig:attn_entropy_conformer}
	\end{subfigure}
	\hfill
	\begin{subfigure}[b]{0.45\textwidth}
		\includegraphics[width=\textwidth]{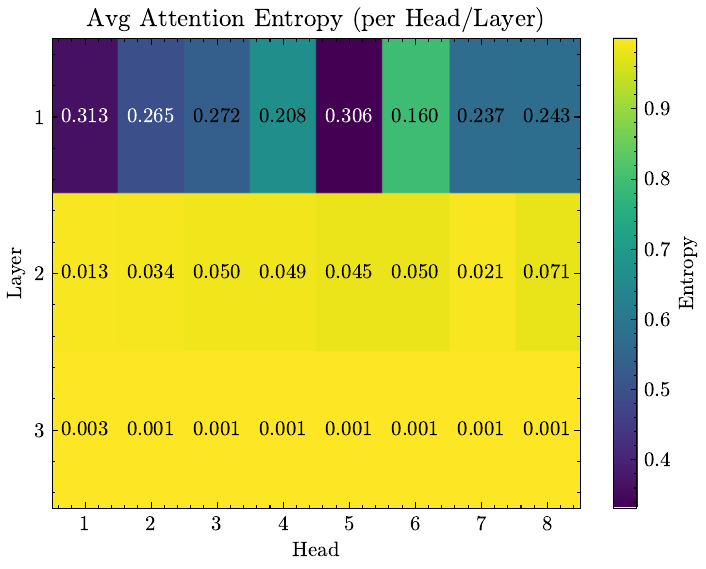}
		\caption{}
		\label{subfig:attn_entropy_transformer}
	\end{subfigure}
	\caption{Normalized attention entropy across all layers and heads, averaged over all test samples. Entropy values were normalized by the maximum possible entropy ($\log(n)$). A value of 0 indicates highly focused attention (one-hot), while 1 reflects fully uniform attention with maximum uncertainty. Numbers within cells indicate the standard deviation across samples; white numbers highlight heads with entropy lower than half of the maximum. \textbf{(a)} HRVConformer maintains diverse and adaptive attention entropy across layers. \textbf{(b)} HRVTransformer exhibits more redundant patterns, with limited head-level diversity.} 
	\label{attn_entropy}
\end{figure}

Attention entropy has been widely employed to interpret, analyze, and improve the performance and representational capacity of attention-based architectures \cite{pardyl2023active, maisonnave2025exploiting, miguelEntropyRegularizedAttentionExplainable2025, jha2025entropy}. \cref{attn_entropy} shows the normalized average attention entropy across all test samples, computed separately for each head and layer. Entropy was calculated using \cref{eqn_Enpy}, where $p_i$ denotes the attention weight of the query on the $i^{th}$ key within a sequence length of $n$. Low entropy indicates sharper attention distributions, where the model concentrates on specific tokens, often reflecting local features. In contrast, high entropy corresponds to more uniform distributions, suggesting reliance on broader, long-range dependencies. Standard deviations are also indicated in each cell, providing insight into variability across samples. Notably, heads with low mean but high variance entropy adapt flexibly across inputs by attending to different tokens, while heads with consistently high entropy and low variance represent diffuse, less informative attention.

\begin{eqnarray}
	\label{eqn_Enpy}
	H(p) = - \sum_{i=1}^{n}p_i log(p_i)
\end{eqnarray}

As shown in \cref{attn_entropy}, the first layer of both models exhibits relatively low mean and high variance entropy, indicating an ability to focus on salient tokens while adapting effectively across different samples. In deeper layers, mean entropy increases while variance decreases, reflecting a shift toward more globally distributed attention. HRVTransformer demonstrates higher mean entropy but lower variance compared with HRVConformer, suggesting that it distributes attention broadly and uniformly (“looks everywhere”), whereas HRVConformer preserves the ability to emphasize specific, longer-range tokens. Moreover, HRVConformer maintains substantial diversity across heads, with some heads showing distinctly higher or lower entropy, whereas HRVTransformer heads converge toward nearly identical entropy values in the final layer, revealing redundancy. The persistent non-negligible variance across HRVConformer heads in the last two layers suggests stronger adaptability and greater capacity to differentiate features across samples, while HRVTransformer’s near-zero variance in the final layer indicates limited representational diversity.

Due to the limited availability of training data, the full potential of the proposed model could not be fully demonstrated. Nevertheless, this architecture holds significant promise for HR signal processing, particularly when scaled with larger and more diverse datasets. Another limitation lies in the labeling process. All strong labels were provided by a single expert, which may introduce a degree of subjective bias. Furthermore, the weak labels were generated based on these strong labels, meaning that any single incorrect strong label could propagate errors across multiple weakly labeled one-hour epochs (typically between five to eleven epochs), compounding the labeling noise and potentially degrading model performance.

This limitation is also reflected in the validation metrics. During training, the validation loss continued to increase even though the validation accuracy remained stable. This discrepancy suggests that the model was becoming increasingly confident in certain predictions that were, in fact, incorrect, indicating a misalignment between the model’s learned representations and the potentially noisy labels. In other words, while the model's confidence was growing, its decisions were sometimes in conflict with the assigned labels, highlighting the importance of high-quality and consistent annotations for model development and evaluation.

\section{Conclusion}

In summary, this study demonstrates that HRVConformer can achieve strong performance in classifying hypoxic–ischemic encephalopathy (HIE) directly from raw heart rate signals with weak supervision, without reliance on hand-crafted features. This hybrid architecture surpass the pure transformer and convolutional architectures, highlighting the advantage of integrating convolution and transformer on capturing local and long-distance dependencies. The analysis and visualization on the learned attention demonstrates the model's reliability and kind of interpretability. The scalability also indicates the potential of the model with more training data. In addition, The enhanced version of Pan-Tompkins algorithm significantly improved the quality and coverage of the extracted heart rate signal from very noisy ECG signal, which pave the way for downstream large scale HRV analysis. Finally, although HRV provides a simpler representation than EEG, its real-time and non-invasive nature makes it a valuable biomarker in NICUs. These findings highlight HRV’s potential to complement existing modalities and contribute to earlier detection, better prognostication, and personalized treatment strategies for infants with HIE.

\bibliographystyle{unsrt}  
\bibliography{references}  

\begin{thebibliography}{10}

\bibitem{hankinsDefiningPathogenesisPathophysiology2003}
G~Hankins.
\newblock Defining the pathogenesis and pathophysiology of neonatal
  encephalopathy and cerebral palsy.
\newblock {\em Obstetrics \& Gynecology}, 102(3):628--636, September 2003.

\bibitem{kurinczukEpidemiologyNeonatalEncephalopathy2010}
Jennifer~J. Kurinczuk, Melanie {White-Koning}, and Nadia Badawi.
\newblock Epidemiology of neonatal encephalopathy and hypoxic--ischaemic
  encephalopathy.
\newblock {\em Early Human Development}, 86(6):329--338, June 2010.

\bibitem{allenHypoxicIschemicEncephalopathy2011}
Kimberly~A. Allen and Debra~H. Brandon.
\newblock Hypoxic {{Ischemic Encephalopathy}}: {{Pathophysiology}} and
  {{Experimental Treatments}}.
\newblock {\em Newborn and Infant Nursing Reviews}, 11(3):125--133, September
  2011.

\bibitem{lawnTwoMillionIntrapartumrelated2009}
Joy~E. Lawn, Anne~CC Lee, Mary Kinney, Lynn Sibley, Wally~A. Carlo, Vinod~K.
  Paul, Robert Pattinson, and Gary~L. Darmstadt.
\newblock Two million intrapartum-related stillbirths and neonatal deaths:
  {{Where}}, why, and what can be done?
\newblock {\em International Journal of Gynecology \& Obstetrics},
  107(Supplement):S5--S19, 2009.

\bibitem{committee2014hypothermia}
Committee on~Fetus and Newborn.
\newblock Hypothermia and neonatal encephalopathy.
\newblock {\em Pediatrics}, 133(6):1146--1150, 2014.

\bibitem{ashooriMachineLearningModels2024}
Minoo Ashoori, John~M. O'Toole, Aisling~A. Garvey, Ken~D. O'Halloran, Brian
  Walsh, Michael Moore, Andreea~M. Pavel, Geraldine~B. Boylan, Deirdre~M.
  Murray, Eugene~M. Dempsey, and Fiona~B. McDonald.
\newblock Machine learning models of cerebral oxygenation ({{rcSO2}}) for brain
  injury detection in neonates with hypoxic-ischaemic encephalopathy.
\newblock {\em The Journal of Physiology}, 602(22):6347--6360, 2024.

\bibitem{kosteczkoTherapeuticHypothermiaForm2025}
Pola Kosteczko, Patrycja~Maria Pelczar, Magdalena Kosanowska, Klaudia
  Ko{\'s}la, and Adrianna Wieleba.
\newblock Therapeutic hypothermia as a form of neonatal hypoxic-ischemic
  encephalopathy neuroprotection and novel therapeutic options.
\newblock {\em Journal of Pre-Clinical and Clinical Research}, September 2025.

\bibitem{vergalesDepressedHeartRate2014}
Brooke~D. Vergales, Santina~A. Zanelli, Julie~A. Matsumoto, Howard~P. Goodkin,
  Douglas~E. Lake, J.~Randall Moorman, and Karen~D. Fairchild.
\newblock Depressed {{Heart Rate Variability}} is {{Associated}} with
  {{Abnormal EEG}}, {{MRI}}, and {{Death}} in {{Neonates}} with {{Hypoxic
  Ischemic Encephalopathy}}.
\newblock {\em American journal of perinatology}, 31(10):855--862, November
  2014.

\bibitem{addisonHeartRateCharacteristics2009}
K~Addison, M~P Griffin, J~R Moorman, D~E Lake, and T~M O'Shea.
\newblock Heart rate characteristics and neurodevelopmental outcome in very low
  birth weight infants.
\newblock {\em Journal of Perinatology}, 29(11):750--756, November 2009.

\bibitem{andersenSeverityHypoxicIschemic2019}
Mads Andersen, Ted C.~K. Andelius, Mette~V. Pedersen, Kasper~J. Kyng, and
  Tine~B. Henriksen.
\newblock Severity of hypoxic ischemic encephalopathy and heart rate
  variability in neonates: A systematic review.
\newblock {\em BMC Pediatrics}, 19(1):242, December 2019.

\bibitem{gouldingHeartRateVariability2017}
Robert~M. Goulding, Nathan~J. Stevenson, Deirdre~M. Murray, Vicki Livingstone,
  Peter~M. Filan, and Geraldine~B. Boylan.
\newblock Heart rate variability in hypoxic ischemic encephalopathy during
  therapeutic hypothermia.
\newblock {\em Pediatric Research}, 81(4):609--615, April 2017.

\bibitem{bersaniHeartRateVariability2021}
Iliana Bersani, Fiammetta Piersigilli, Diego Gazzolo, Francesca Campi,
  Immacolata Savarese, Andrea Dotta, Pietro~Paolo Tamborrino, Cinzia Auriti,
  and Corrado Di~Mambro.
\newblock Heart rate variability as possible marker of brain damage in neonates
  with hypoxic ischemic encephalopathy: A systematic review.
\newblock {\em European Journal of Pediatrics}, 180(5):1335--1345, May 2021.

\bibitem{massaroEffectTemperatureHeart2017}
An~N. Massaro, Heather~E. Campbell, Marina Metzler, Tareq {Al-Shargabi}, Yunfei
  Wang, Adre {du Plessis}, and R.B. Govindan.
\newblock Effect of temperature on heart rate variability in neonatal intensive
  care unit patients with hypoxic ischemic encephalopathy.
\newblock {\em Pediatric critical care medicine : a journal of the Society of
  Critical Care Medicine and the World Federation of Pediatric Intensive and
  Critical Care Societies}, 18(4):349--354, April 2017.

\bibitem{edoigiawerieSystematicReviewEEG2024}
Sylvia Edoigiawerie, Julia Henry, Naoum Issa, and Henry David.
\newblock A {{Systematic Review}} of {{EEG}} and {{MRI Features}} for
  {{Predicting Long-Term Neurological Outcomes}} in {{Cooled Neonates With
  Hypoxic-Ischemic Encephalopathy}} ({{HIE}}).
\newblock {\em Cureus}, October 2024.

\bibitem{farihaAnalysisPanTompkinsAlgorithm2020}
M.~A.~Z. Fariha, R.~Ikeura, S.~Hayakawa, and S.~Tsutsumi.
\newblock Analysis of {{Pan-Tompkins Algorithm Performance}} with {{Noisy ECG
  Signals}}.
\newblock {\em Journal of Physics: Conference Series}, 1532(1):012022, June
  2020.

\bibitem{imtiazPanTompkinsRobustApproach2022}
Md~Niaz Imtiaz and Naimul Khan.
\newblock Pan-{{Tompkins}}++: {{A Robust Approach}} to {{Detect R-peaks}} in
  {{ECG Signals}}.
\newblock In {\em 2022 {{IEEE International Conference}} on {{Bioinformatics}}
  and {{Biomedicine}} ({{BIBM}})}, pages 2905--2912, December 2022.

\bibitem{hamiltonQuantitativeInvestigationQRS1986}
Patrick~S. Hamilton and Willis~J. Tompkins.
\newblock Quantitative {{Investigation}} of {{QRS Detection Rules Using}} the
  {{MIT}}/{{BIH Arrhythmia Database}}.
\newblock {\em IEEE Transactions on Biomedical Engineering},
  BME-33(12):1157--1165, December 1986.

\bibitem{ranaCardiacDiseaseDetection2019}
Amrita Rana and Kyung~Ki Kim.
\newblock Cardiac {{Disease Detection Using Modified Pan-Tompkins Algorithm}}.
\newblock {\em Journal of Sensor Science and Technology}, 28(1):13--16, January
  2019.

\bibitem{esgalhadoPeakDetectionHRV2022}
Filipa Esgalhado, Arnaldo Batista, Valentina Vassilenko, Sara Russo, and Manuel
  Ortigueira.
\newblock Peak {{Detection}} and {{HRV Feature Evaluation}} on {{ECG}} and
  {{PPG Signals}}.
\newblock {\em Symmetry}, 14(6):1139, 2022.

\bibitem{sahambiNewApproachOnline1996}
J.S. Sahambi, S.N. Tandon, and R.K.P. Bhatt.
\newblock A new approach for on-line {{ECG}} characterization.
\newblock In {\em Proceedings of the 1996 {{Fifteenth Southern Biomedical
  Engineering Conference}}}, pages 409--411, March 1996.

\bibitem{goyalStudyHRVDynamics2012}
Yamini Goyal and Anshul Jain.
\newblock Study of {{HRV Dynamics}} and {{Comparison Using Wavelet Analysis}}
  and {{Pan Tompkins Algorithm}}.
\newblock In {\em 2012 {{ASE}}/{{IEEE International Conference}} on
  {{BioMedical Computing}} ({{BioMedCom}})}, pages 43--49, December 2012.

\bibitem{gautamFeatureExtractionHRV2017}
Desh~Deepak Gautam, V.~K. Giri, and K.~G. Upadhyay.
\newblock Feature extraction of {{HRV}} signal using wavelet transform.
\newblock In {\em 2017 2nd {{International Conference}} for {{Convergence}} in
  {{Technology}} ({{I2CT}})}, pages 1030--1034, April 2017.

\bibitem{vermaRobustAlgorithmDerivation2013}
A.~Verma, S.~Cabrera, A.~Mayorga, and H.~Nazeran.
\newblock A {{Robust Algorithm}} for {{Derivation}} of {{Heart Rate Variability
  Spectra}} from {{ECG}} and {{PPG Signals}}.
\newblock In {\em 2013 29th {{Southern Biomedical Engineering Conference}}},
  pages 35--36, Miami, FL, May 2013. IEEE.

\bibitem{martinezApplicationPhasorTransform2010}
Arturo Mart{\'i}nez, Ra{\'u}l Alcaraz, and Jos{\'e}~Joaqu{\'i}n Rieta.
\newblock Application of the phasor transform for automatic delineation of
  single-lead {{ECG}} fiducial points.
\newblock {\em Physiological Measurement}, 31(11):1467--1485, November 2010.

\bibitem{pavelHeartRateVariability2023}
Andreea~M. Pavel, Sean~R. Mathieson, Vicki Livingstone, John~M. O'Toole,
  Ronit~M. Pressler, Linda~S. de~Vries, Janet~M. Rennie, Subhabrata Mitra,
  Eugene~M. Dempsey, Deirdre~M. Murray, William~P. Marnane, Geraldine~B.
  Boylan, and {\relax Ans}eR Consortium.
\newblock Heart rate variability analysis for the prediction of {{EEG}} grade
  in infants with hypoxic ischaemic encephalopathy within the first 12 h of
  birth.
\newblock {\em Frontiers in Pediatrics}, 10, January 2023.

\bibitem{metzlerPatternBrainInjury2017}
Marina Metzler, Rathinaswamy Govindan, Tareq {Al-Shargabi}, Gilbert Vezina,
  Nickie Andescavage, Yunfei Wang, Adre Du~Plessis, and An~N Massaro.
\newblock Pattern of brain injury and depressed heart rate variability in
  newborns with hypoxic ischemic encephalopathy.
\newblock {\em Pediatric Research}, 82(3):438--443, September 2017.

\bibitem{ersoyFeatureExtractionBased2016}
Ersin Ersoy and Mahmut Hekim.
\newblock Feature {{Extraction Based}} on {{Pan Tompkins Algorithm}} from {{ECG
  Signals}} and {{Diagnosis}} of {{Arrhythmia Using Multilayer Perceptron
  Neural Network}}.
\newblock {\em Journal of New Results in Science}, 5, November 2016.

\bibitem{turnipDetectionPotentialHypertension2024}
Mardi Turnip, Abdi Dharma, Siti Aisyah, Delima Sitanggang, {Yennimar}, Mohammad
  Taufik, Dhanny~Rukmana Manday, and Arjon Turnip.
\newblock Detection of {{Potential Hypertension}} with {{Pan Tompkins
  Extraction}} and {{Naive Bayes Classifier Methods}}.
\newblock In {\em 2024 {{IEEE International Conference}} on {{Artificial
  Intelligence}} and {{Mechatronics Systems}} ({{AIMS}})}, pages 1--7, February
  2024.

\bibitem{semenovaPredictionShorttermHealth2019}
Oksana Semenova, Giorgia Carra, Gordon Lightbody, Geraldine Boylan, Eugene
  Dempsey, and Andriy Temko.
\newblock Prediction of short-term health outcomes in preterm neonates from
  heart-rate variability and blood pressure using boosted decision trees.
\newblock {\em Computer Methods and Programs in Biomedicine}, 180:104996,
  October 2019.

\bibitem{yasovabarbeauHeartRateVariability2019}
Daphna Yasova~Barbeau, Charlene Krueger, Melissa Huene, Nicole Copenhaver,
  Jeffrey Bennett, Michael Weaver, and Michael~D. Weiss.
\newblock Heart rate variability and inflammatory markers in neonates with
  hypoxic-ischemic encephalopathy.
\newblock {\em Physiological Reports}, 7(15):e14110, 2019.

\bibitem{al-shargabiInflammatoryCytokineResponse2017}
T~{Al-Shargabi}, R~B Govindan, R~Dave, M~Metzler, Y~Wang, A~Du~Plessis, and A~N
  Massaro.
\newblock Inflammatory cytokine response and reduced heart rate variability in
  newborns with hypoxic-ischemic encephalopathy.
\newblock {\em Journal of Perinatology}, 37(6):668--672, June 2017.

\bibitem{wangECGStressDetection2023}
Ling Wang, Jiayu Hao, Tie~Hua Zhou, and Fangjie Song.
\newblock {{ECG Stress Detection Model Based}} on {{Heart Rate Variability
  Feature Extraction}}.
\newblock In {\em Proceedings of the 2023 7th {{International Conference}} on
  {{High Performance Compilation}}, {{Computing}} and {{Communications}}},
  pages 184--188, Jinan China, June 2023. ACM.

\bibitem{sharanECGDerivedHeartRate2020}
Roneel~V. Sharan, Shlomo Berkovsky, Hao Xiong, and Enrico Coiera.
\newblock {{ECG-Derived Heart Rate Variability Interpolation}} and 1-{{D
  Convolutional Neural Networks}} for {{Detecting Sleep Apnea}}.
\newblock In {\em 2020 42nd {{Annual International Conference}} of the {{IEEE
  Engineering}} in {{Medicine}} \& {{Biology Society}} ({{EMBC}})}, pages
  637--640, July 2020.

\bibitem{gouldingHeartRateVariability2015}
Robert~M. Goulding, Nathan~J. Stevenson, Deirdre~M. Murray, Vicki Livingstone,
  Peter~M. Filan, and Geraldine~B. Boylan.
\newblock Heart rate variability in hypoxic ischemic encephalopathy:
  Correlation with {{EEG}} grade and 2-y neurodevelopmental outcome.
\newblock {\em Pediatric Research}, 77(5):681--687, May 2015.

\bibitem{aliefendiogluHeartRateVariability2012}
Didem Aliefendio{\u g}lu, Tolga Do{\u g}ru, Meryem Albayrak, Emine
  DibekM{\i}s{\i}rl{\i}o{\u g}lu, and Cihat {\c S}anl{\i}.
\newblock Heart {{Rate Variability}} in {{Neonates}} with {{Hypoxic Ischemic
  Encephalopathy}}.
\newblock {\em The Indian Journal of Pediatrics}, 79(11):1468--1472, November
  2012.

\bibitem{osathitpornRRWaveNetCompactEndtoEnd2023}
Pongpanut Osathitporn, Guntitat Sawadwuthikul, Punnawish Thuwajit, Kawisara
  Ueafuea, Thee Mateepithaktham, Narin Kunaseth, Tanut Choksatchawathi,
  Proadpran Punyabukkana, Emmanuel Mignot, and Theerawit Wilaiprasitporn.
\newblock {{RRWaveNet}}: {{A Compact End-to-End Multiscale Residual CNN}} for
  {{Robust PPG Respiratory Rate Estimation}}.
\newblock {\em IEEE Internet of Things Journal}, 10(18):15943--15952, September
  2023.

\bibitem{songHeartRateEstimation2020}
Rencheng Song, Senle Zhang, Chang Li, Yunfei Zhang, Juan Cheng, and Xun Chen.
\newblock Heart {{Rate Estimation From Facial Videos Using}} a {{Spatiotemporal
  Representation With Convolutional Neural Networks}}.
\newblock {\em IEEE Transactions on Instrumentation and Measurement},
  69(10):7411--7421, October 2020.

\bibitem{karhadeTimeFrequencyDomain2022}
Jay Karhade, Shaswati Dash, Samit~Kumar Ghosh, Dinesh~Kumar Dash, and
  Rajesh~Kumar Tripathy.
\newblock Time--{{Frequency-Domain Deep Learning Framework}} for the
  {{Automated Detection}} of {{Heart Valve Disorders Using PCG Signals}}.
\newblock {\em IEEE Transactions on Instrumentation and Measurement}, 71:1--11,
  2022.

\bibitem{ramalakshmiDevelopingDeepLearning2024}
V.~Ramalakshmi, Neha Garg, Manzoore~Elahi M.~Soudagar, Daxa Vekariya, Harshal
  Patil, and Arun Shalin~L V.
\newblock Developing {{Deep Learning Model}} for {{Abnormal Heartbeat
  Detection}} in {{Electrocardiography Signals}}.
\newblock In {\em 2024 2nd {{International Conference}} on {{Intelligent Data
  Communication Technologies}} and {{Internet}} of {{Things}} ({{IDCIoT}})},
  pages 315--320, January 2024.

\bibitem{he2016deep}
Kaiming He, Xiangyu Zhang, Shaoqing Ren, and Jian Sun.
\newblock Deep residual learning for image recognition.
\newblock In {\em Proceedings of the IEEE conference on computer vision and
  pattern recognition}, pages 770--778, 2016.

\bibitem{krizhevsky2012imagenet}
Alex Krizhevsky, Ilya Sutskever, and Geoffrey~E Hinton.
\newblock Imagenet classification with deep convolutional neural networks.
\newblock {\em Advances in neural information processing systems}, 25, 2012.

\bibitem{dosovitskiyImageWorth16x162021}
Alexey Dosovitskiy, Lucas Beyer, Alexander Kolesnikov, Dirk Weissenborn,
  Xiaohua Zhai, Thomas Unterthiner, Mostafa Dehghani, Matthias Minderer, Georg
  Heigold, Sylvain Gelly, Jakob Uszkoreit, and Neil Houlsby.
\newblock An {{Image}} is {{Worth}} 16x16 {{Words}}: {{Transformers}} for
  {{Image Recognition}} at {{Scale}}, June 2021.

\bibitem{wang2024medformer}
Yihe Wang, Nan Huang, Taida Li, Yujun Yan, and Xiang Zhang.
\newblock Medformer: A multi-granularity patching transformer for medical
  time-series classification.
\newblock {\em Advances in Neural Information Processing Systems},
  37:36314--36341, 2024.

\bibitem{anwarTransformersBiosignalAnalysis2025}
Ayman Anwar, Yassin Khalifa, James~L. Coyle, and Ervin Sejdic.
\newblock Transformers in biosignal analysis: {{A}} review.
\newblock {\em Information Fusion}, 114:102697, February 2025.

\bibitem{pengConformerLocalFeatures2021}
Zhiliang Peng, Wei Huang, Shanzhi Gu, Lingxi Xie, Yaowei Wang, Jianbin Jiao,
  and Qixiang Ye.
\newblock Conformer: {{Local Features Coupling Global Representations}} for
  {{Visual Recognition}}.
\newblock In {\em 2021 {{IEEE}}/{{CVF International Conference}} on {{Computer
  Vision}} ({{ICCV}})}, pages 357--366, Montreal, QC, Canada, October 2021.
  IEEE.

\bibitem{fuUformerUnetBased2022}
Yihui Fu, Yun Liu, Jingdong Li, Dawei Luo, Shubo Lv, Yukai Jv, and Lei Xie.
\newblock Uformer: {{A Unet Based Dilated Complex}} \& {{Real Dual-Path
  Conformer Network}} for {{Simultaneous Speech Enhancement}} and
  {{Dereverberation}}.
\newblock In {\em {{ICASSP}} 2022 - 2022 {{IEEE International Conference}} on
  {{Acoustics}}, {{Speech}} and {{Signal Processing}} ({{ICASSP}})}, pages
  7417--7421, May 2022.

\bibitem{songEEGConformerConvolutional2023}
Yonghao Song, Qingqing Zheng, Bingchuan Liu, and Xiaorong Gao.
\newblock {{EEG Conformer}}: {{Convolutional Transformer}} for {{EEG Decoding}}
  and {{Visualization}}.
\newblock {\em IEEE Transactions on Neural Systems and Rehabilitation
  Engineering}, 31:710--719, 2023.

\bibitem{liCNNInformerHybridDeep2025}
Chuanyu Li, Haotian Li, Xingchen Dong, Xiangwen Zhong, Haozhou Cui, Dezan Ji,
  Landi He, Guoyang Liu, and Weidong Zhou.
\newblock {{CNN-Informer}}: {{A}} hybrid deep learning model for seizure
  detection on long-term {{EEG}}.
\newblock {\em Neural Networks}, 181:106855, January 2025.

\bibitem{gulatiConformerConvolutionaugmentedTransformer2020}
Anmol Gulati, James Qin, Chung-Cheng Chiu, Niki Parmar, Yu~Zhang, Jiahui Yu,
  Wei Han, Shibo Wang, Zhengdong Zhang, Yonghui Wu, and Ruoming Pang.
\newblock Conformer: {{Convolution-augmented Transformer}} for {{Speech
  Recognition}}, May 2020.

\bibitem{rekeshFastConformerLinearly2023}
Dima Rekesh, Nithin~Rao Koluguri, Samuel Kriman, Somshubra Majumdar, Vahid
  Noroozi, He~Huang, Oleksii Hrinchuk, Krishna Puvvada, Ankur Kumar, Jagadeesh
  Balam, and Boris Ginsburg.
\newblock Fast {{Conformer With Linearly Scalable Attention For Efficient
  Speech Recognition}}.
\newblock In {\em 2023 {{IEEE Automatic Speech Recognition}} and
  {{Understanding Workshop}} ({{ASRU}})}, pages 1--8, December 2023.

\bibitem{kimSEConformerTimeDomainSpeech2021}
Eesung Kim and Hyeji Seo.
\newblock {{SE-Conformer}}: {{Time-Domain Speech Enhancement Using Conformer}}.
\newblock In {\em Interspeech 2021}, pages 2736--2740. ISCA, August 2021.

\bibitem{rennieCharacterisationNeonatalSeizures2019}
Janet~M Rennie, Linda~S De~Vries, Mats Blennow, Adrienne Foran, Divyen~K Shah,
  Vicki Livingstone, Alexander~C Van~Huffelen, Sean~R Mathieson, Elena
  Pavlidis, Lauren~C Weeke, Mona~C Toet, Mikael Finder, Raga~Mallika
  Pinnamaneni, Deirdre~M Murray, Anthony~C Ryan, William~P Marnane, and
  Geraldine~B Boylan.
\newblock Characterisation of neonatal seizures and their treatment using
  continuous {{EEG}} monitoring: A multicentre experience.
\newblock {\em Archives of Disease in Childhood - Fetal and Neonatal Edition},
  104(5):F493--F501, September 2019.

\bibitem{pavelMachinelearningAlgorithmNeonatal2020}
Andreea~M Pavel, Janet~M Rennie, Linda~S De~Vries, Mats Blennow, Adrienne
  Foran, Divyen~K Shah, Ronit~M Pressler, Olga Kapellou, Eugene~M Dempsey,
  Sean~R Mathieson, Elena Pavlidis, Alexander~C Van~Huffelen, Vicki
  Livingstone, Mona~C Toet, Lauren~C Weeke, Mikael Finder, Subhabrata Mitra,
  Deirdre~M Murray, William~P Marnane, and Geraldine~B Boylan.
\newblock A machine-learning algorithm for neonatal seizure recognition: A
  multicentre, randomised, controlled trial.
\newblock {\em The Lancet Child \& Adolescent Health}, 4(10):740--749, October
  2020.

\bibitem{murrayEarlyEEGFindings2009}
Deirdre~M. Murray, Geraldine~B. Boylan, Cornelius~A. Ryan, and Sean Connolly.
\newblock Early {{EEG Findings}} in {{Hypoxic-Ischemic Encephalopathy Predict
  Outcomes}} at 2 {{Years}}.
\newblock {\em Pediatrics}, 124(3):e459--e467, September 2009.

\bibitem{panRealTimeQRSDetection1985}
Jiapu Pan and Willis~J. Tompkins.
\newblock A {{Real-Time QRS Detection Algorithm}}.
\newblock {\em IEEE Transactions on Biomedical Engineering},
  BME-32(3):230--236, March 1985.

\bibitem{alvarezComparisonThreeQRS2013}
Ra{\'u}l~Alonso {\'A}lvarez, Arturo J.~M{\'e}ndez Pen{\'i}n, and
  X.~Ant{\'o}n~Vila Sobrino.
\newblock A {{Comparison}} of {{Three QRS Detection Algorithms Over}} a
  {{Public Database}}.
\newblock {\em Procedia Technology}, 9:1159--1165, 2013.

\bibitem{vestOpenSourceBenchmarked2018}
Adriana~N Vest, Giulia Da~Poian, Qiao Li, Chengyu Liu, Shamim Nemati, Amit~J
  Shah, and Gari~D Clifford.
\newblock An open source benchmarked toolbox for cardiovascular waveform and
  interval analysis.
\newblock {\em Physiological Measurement}, 39(10):105004, October 2018.

\bibitem{shaffer2017overview}
Fred Shaffer and Jay~P Ginsberg.
\newblock An overview of heart rate variability metrics and norms.
\newblock {\em Frontiers in public health}, 5:258, 2017.

\bibitem{voss2015short}
Andreas Voss, Rico Schroeder, Andreas Heitmann, Annette Peters, and Siegfried
  Perz.
\newblock Short-term heart rate variability—influence of gender and age in
  healthy subjects.
\newblock {\em PloS one}, 10(3):e0118308, 2015.

\bibitem{adam2023heart}
Josephine Adam, Sven Rupprecht, Erika~CS K{\"u}nstler, and Dirk Hoyer.
\newblock Heart rate variability as a marker and predictor of inflammation,
  nosocomial infection, and sepsis--a systematic review.
\newblock {\em Autonomic Neuroscience}, 249:103116, 2023.

\bibitem{daiTransformerXLAttentiveLanguage2019}
Zihang Dai, Zhilin Yang, Yiming Yang, Jaime Carbonell, Quoc Le, and Ruslan
  Salakhutdinov.
\newblock Transformer-{{XL}}: {{Attentive Language Models}} beyond a
  {{Fixed-Length Context}}.
\newblock In {\em Proceedings of the 57th {{Annual Meeting}} of the
  {{Association}} for {{Computational Linguistics}}}, pages 2978--2988,
  Florence, Italy, 2019. Association for Computational Linguistics.

\bibitem{rezaeiAssessingEffectivenessHeart2024a}
K.~Rezaei, K.~Yu, S.~R. Mathieson, A.~Flynn, G.~Lightbody, G.B. Boylan, and
  W.~P. Marnane.
\newblock Assessing the {{Effectiveness}} of {{Heart Rate Variability}} as {{A
  Diagnostic Tool}} for {{Brain Injuries}} in {{Infants}}.
\newblock In {\em 2024 46th {{Annual International Conference}} of the {{IEEE
  Engineering}} in {{Medicine}} and {{Biology Society}} ({{EMBC}})}, pages
  1--4, July 2024.

\bibitem{yu2023neonatal}
Shuwen Yu, William~P Marnane, Geraldine~B Boylan, and Gordon Lightbody.
\newblock Neonatal hypoxic-ischemic encephalopathy grading from multi-channel
  eeg time-series data using a fully convolutional neural network.
\newblock {\em Technologies}, 11(6):151, 2023.

\bibitem{abnar2020quantifying}
Samira Abnar and Willem Zuidema.
\newblock Quantifying attention flow in transformers.
\newblock {\em arXiv preprint arXiv:2005.00928}, 2020.

\bibitem{pardyl2023active}
Adam Pardyl, Grzegorz Rype{\'s}{\'c}, Grzegorz Kurzejamski, Bartosz
  Zieli{\'n}ski, and Tomasz Trzci{\'n}ski.
\newblock Active visual exploration based on attention-map entropy.
\newblock {\em arXiv preprint arXiv:2303.06457}, 2023.

\bibitem{maisonnave2025exploiting}
Lucas Maisonnave, Karim Haroun, and Tom P{\'e}geot.
\newblock Exploiting information redundancy in attention maps for extreme
  quantization of vision transformers.
\newblock In {\em Proceedings of the IEEE/CVF International Conference on
  Computer Vision}, pages 4033--4041, 2025.

\bibitem{miguelEntropyRegularizedAttentionExplainable2025}
Pedro~L. Miguel, Leandro~A. Neves, Alessandra Lumini, Giuliano~C. Medalha,
  Guilherme~F. Roberto, Guilherme~B. Rozendo, Adriano~M. Cansian, Tha{\'i}na
  A.~A. Tosta, and Marcelo~Z. {do Nascimento}.
\newblock Entropy-{{Regularized Attention}} for {{Explainable Histological
  Classification}} with {{Convolutional}} and {{Hybrid Models}}.
\newblock {\em Entropy}, 27(7):722, July 2025.

\bibitem{jha2025entropy}
Nandan~Kumar Jha and Brandon Reagen.
\newblock Entropy-guided attention for private llms.
\newblock {\em arXiv preprint arXiv:2501.03489}, 2025.

\end{thebibliography}




\end{document}